%% file: paper.tex
\documentclass{article}

\usepackage[final, nonatbib]{neurips_2021}

\usepackage[utf8]{inputenc} 
\usepackage[T1]{fontenc}    
\usepackage{hyperref}       
\usepackage{url}            
\usepackage{booktabs}       
\usepackage{amsfonts}       
\usepackage{nicefrac}       
\usepackage{microtype}      
\usepackage{xcolor}         
\usepackage{soul}
\usepackage{multirow}

\title{Snowflake: Scaling GNNs to high-dimensional continuous control via parameter freezing}

\author{%
  Charlie Blake\thanks{Corresponding author. Now at Graphcore, Bristol}\\
  University of Oxford\\
  \texttt{thecharlieblake@gmail.com}\\
  \And
  Vitaly Kurin\\
  University of Oxford\\
  \texttt{vitaly.kurin@cs.ox.ac.uk}\\
  \And
  Maximilian Igl\thanks{Now at Waymo, Oxford}\\
  University of Oxford\\
  \texttt{maximilian.igl@gmail.com}\\
  \And
  Shimon Whiteson\\
  University of Oxford\\
  \texttt{shimon.whiteson@cs.ox.ac.uk}\\
}

\usepackage[acronym,smallcaps,nowarn,section,nogroupskip,nonumberlist,nohypertypes={acronym,notation}]{glossaries}
\usepackage[numbers]{natbib}
\usepackage[pdftex]{graphicx}
\usepackage{subcaption}
\usepackage{wrapfig}

\setacronymstyle{long-sc-short}

\newcommand{\Tau}{\mathrm{T}}

\newacronym{sf}{Snowflake}{Snowflake}
\newacronym{gnn}{GNN}{graph neural network}
\newacronym{mlp}{MLP}{multi-layer perceptron}
\newacronym{nn}{NerveNet}{NerveNet}
\newacronym{rl}{RL}{reinforcement learning}
\newacronym{mpnn}{MPNN}{message passing neural network}
\newacronym{ppo}{PPO}{proximal policy optimisation}
\newacronym{mdp}{MDP}{Markov decision process}
\newacronym{gru}{GRU}{gated recurrent unit}
\newacronym{pg}{PG}{Policy Gradients}
\newacronym{smp}{SMP}{Shared Modular Policies}
\newacronym{dgn}{DGN}{dynamic graph networks}
\newacronym{am}{Amorpheus}{Amorpheus}
\DeclareMathOperator{\E}{\mathbb{E}}

\usepackage{xcolor}

\newcommand{\mi}[1]{\textcolor{black}{#1}}

\newcommand{\cb}[1]{\textcolor{black}{#1}}
\newcommand{\cam}[1]{\textcolor{black}{#1}}
\newcommand{\cbb}[1]{\textcolor{black}{#1}}

\begin{document}

\maketitle

\begin{abstract}
\input{sections/Abstract}
\end{abstract}

\vspace{-0.5em}
\section{Introduction}
\label{Introduction}
\input{sections/Introduction}

\vspace{-0.5em}
\section{Background}
\label{Background}
\input{sections/Background}

\vspace{-0.5em}
\section{Analysing GNN Scaling Challenges}
\label{Scaling}
\input{sections/Scaling}

\vspace{-0.5em}
\section{Related Work}
\label{Related Work}
\input{sections/Related_Work}

\vspace{-0.5em}
\section{Experiments}
\label{Experiments}
\input{sections/Experiments}

\vspace{-0.5em}
\section{Conclusion}
\label{Conclusion}
\input{sections/Conclusion}

\begin{ack}
\cam{VK is a doctoral student at the University of Oxford funded by Samsung R\&D Institute UK through the AIMS program. SW has received funding from the European Research Council under the European Union's Horizon 2020 research and innovation programme (grant agreement number 637713). The experiments were made possible by a generous equipment grant from NVIDIA.}
\end{ack}

\clearpage
\bibliographystyle{plainnat}
\bibliography{paper}


\newpage
\appendix

\section{Appendix}
\input{sections/Appendix}

\end{document}

%% file: sections/Abstract.tex
Recent research has shown that \glspl{gnn} can learn policies for locomotion control that are as effective as a typical \gls{mlp}, with superior transfer and multi-task performance \citep{wang2018nervenet, huang2020one}.
However, results have so far been limited to training on small agents, with the performance of \acrshort{gnn}s deteriorating rapidly as the number of sensors and actuators grows.
A key motivation for the use of \acrshort{gnn}s in the supervised learning setting is their applicability to large graphs, but this benefit has not yet been realised for locomotion control.
We show that poor scaling in \acrshort{gnn}s is a result of increasingly unstable policy updates, caused by overfitting in parts of the network during training.
To combat this, we introduce \acrshort{sf}, a GNN training method for high-dimensional continuous control that freezes parameters in selected parts of the network.
\acrshort{sf} significantly boosts the performance of \acrshort{gnn}s for locomotion control on large agents, now matching the performance of \acrshort{mlp}s while offering superior transfer properties.

%% file: sections/Introduction.tex
Whereas many traditional machine learning models operate on sequential or Euclidean (grid-like) data representations, \glspl{gnn} allow for graph-structured inputs.
\glspl{gnn} have yielded breakthroughs in a variety of complex domains, including drug discovery \citep{lim2019predicting, stokes2020deep}, fraud detection \cite{wang2020apan}, computer vision \citep{shen2018person, sarlin2020superglue}, and particle physics \citep{heintz2020accelerated}.

\Glspl{gnn} have also been successfully applied to \gls{rl}, with promising results on locomotion control tasks with small state and action spaces.
\cb{Not only are \gls{gnn} policies as effective as \glspl{mlp} on certain training tasks, but when a trained policy is transferred to another similar task, \glspl{gnn} significantly outperform \glspl{mlp} \citep{wang2018nervenet,huang2020one}.}
\cb{This is largely due to} the capacity of a single \acrshort{gnn} to operate over arbitrary graph topologies (patterns of connectivity between nodes) and sizes without modification.
\cb{However, so far \glspl{gnn} in \gls{rl} have only shown competitive performance with \glspl{mlp} on lower-dimensional locomotion control tasks.
For higher-dimensional tasks, one must therefore choose between superior training task performance (\glspl{mlp}) and superior transfer performance (\glspl{gnn}).}

This paper investigates the factors underlying poor GNN scaling and introduces a method to combat them.
We begin with an analysis of the \acrshort{gnn}-based \acrshort{nn} architecture \citep{wang2018nervenet}, which we choose for its strong zero-shot transfer performance.
\cbb{We show that optimisation updates for the \gls{gnn} policy have a tendency to cause excessive changes in policy space, leading to performance degrading.}
\cbb{To combat this, current state-of-the-art algorithms \citep{schulman2015trust, schulman2017proximal, abdolmaleki2018maximum} employ trust region-like constraints, inspired by natural gradients \citep{amari1997neural, kakade2001natural}, that limit the change in policy for each update.}
We outline how this policy instability can be framed as a form of overfitting---a problem \gls{gnn} architectures like \acrshort{nn} are known to suffer from in supervised learning, and show that parameter regularisation (a standard remedy for overfitting) leads to a small improvement in \acrshort{nn} performance.

\cbb{We then investigate which structures in the \gls{gnn} contribute most to this overfitting, by applying different learning rates to different parts of the network.
Surprisingly, the best performance is attained when training with a learning rate of zero in the parts of the GNN architecture that encode, decode, and propagate messages in the graph\cb{, in effect training only the part that updates node representations}.}

We use this approach as the basis of our method, \acrshort{sf}, which freezes the parameters of particular operations within the \gls{gnn} to their initialised values, keeping them fixed throughout training while updating the non-frozen parameters as before. This simple technique enables \gls{gnn} policies to be trained much more effectively in high-dimensional environments.


\mi{Experimentally, we show that applying \acrshort{sf} to \acrshort{nn} dramatically improves asymptotic performance and sample complexity on such tasks.}
\mi{We also demonstrate that a policy trained using \acrshort{sf} exhibits improved zero-shot transfer compared to regular \acrshort{nn} or \acrshort{mlp}s on high-dimensional tasks.
}

%% file: sections/Background.tex
\subsection{Reinforcement Learning}
\label{Reinforcement Learning}

We formalise an~\gls{rl} problem as a~\gls{mdp}.
An~\gls{mdp} is a tuple $\langle \mathcal{S}, \mathcal{A}, \mathcal{R}, \mathcal{T}, \rho_0 \rangle$.
The first two elements define the state space $\mathcal{S}$ and the action space $\mathcal{A}$.
At every time step $t$, the agent employs a policy $\pi(a_t|s_t)$ to output a distribution over actions, selects action $a_t\sim\pi(\cdot|s_t)$, and transitions from state $s_t\in\mathcal{S}$ to $s_{t+1}\in\mathcal{S}$, as specified by the transition function $\mathcal{T}(s_{t+1}|s_t,a_t)$ which defines a probability distribution over states.
For the transition, the agent gets a reward $r_t=\mathcal{R}(s_t,a_t,s_{t+1})$.
The last element of an~\gls{mdp} specifies initial distribution over states, i.e., states an agent can be in at time step zero.

Solving an~\gls{mdp} means finding a policy $\pi^*$ that maximises an objective, in our case the expected discounted sum of rewards $J=\E_{\pi}\left[\sum_{t=0}^{\infty}{\gamma^t r_t}\right]$, where $\gamma \in[0,1)$ is a discount factor.
\Gls{pg} \citep{DBLP:journals/corr/ThomasB17} find an optimal policy $\pi^*$ by doing gradient ascent on the objective: $\theta_{t+1} = \theta_t + \alpha \nabla_{\theta}J|_{\theta=\theta_t}$ with $\theta$ parameterising the policy.

Often, to reduce the variance of the gradient estimate, one learns a value function $V(s)=\E_{\pi}\left[\sum_{t=0}^{\infty}\gamma^t r_t \mid s_0=s\right]$, and uses it as a critic of the policy.
In the resulting actor-critic method, the policy gradient takes the form: $\nabla_{\theta}J(\theta) = \E_{\pi_\theta}\left[\sum_tA_t^{\pi_\theta}\nabla_\theta\log{\pi_\theta(a_t|s_t)}\right]$, where $A_t^{\pi_\theta}$
\cbb{is an estimate of the advantage function  $A_t^\pi = \E_{\pi}\left[\sum_{t=0}^{\infty}\gamma^t r_t \mid a_t, s_t\right] - \E_{\pi}\left[\sum_{t=0}^{\infty}\gamma^t r_t \mid s_t \right]$~\citep{schulman2015high}.}

\subsection{Proximal Policy Optimisation}
\label{ppo}

\Gls{ppo} \citep{schulman2015high} is an actor-critic method that has proved effective for a variety of domains including locomotion control \citep{heess2017emergence}.
\gls{ppo} approximates the natural gradient using a first order method, which has the effect of keeping policy updates within a ``trust region''.
This is done through the introduction of a \textit{surrogate objective} to be optimised:
\begin{equation}
\begin{split}
J = \E_{\pi_{\theta'}}\bigg[ \min\bigg(
\frac{\pi_{\theta}(a|s)}{\pi_{\theta'}(a|s)}  A^{\pi_{\theta'}}(s,a),
\text{clip}\Big(\frac{\pi_{\theta}(a|s)}{\pi_{\theta'}(a|s)}, 1 - \epsilon, 1+\epsilon \Big) A^{\pi_{\theta'}}(s,a)\bigg)\bigg]
\end{split}
\end{equation}
where $\epsilon$ is a clipping hyperparameter that effectively limits how much a state-action pair can cause the overall policy to change at each update.
This objective is computed over a number of optimisation epochs, each of which gives an update to the new policy $\pi_{\theta}$.
If during this process a state-action pair with a positive advantage $A^{\pi_{\theta'}}(s,a)$ reaches the upper clipping boundary, the objective no longer provides an incentive for the policy to be improved with respect to that data point.
This similarly applies to state-action pairs with a negative advantage if the lower clipping limit is reached.

\subsection{Graph Neural Networks}
\label{Graph Neural Networks}

\acrshort{gnn}s are a class of neural architecture designed to operate over graph-structured data.
We define a graph as a tuple $\mathcal{G} = (V, E)$ comprising a set of nodes $V$ and edges $E = \{(u, v) \mid u, v \in V\} $.
A labelled graph has corresponding feature vectors for each node and edge that form a pair of matrices $\mathcal{L}_\mathcal{G} = (\boldsymbol{V}, \boldsymbol{E})$, where $\boldsymbol{V} = \{\mathbf{v}_v \in \mathbb{R}^p \mid v \in V\}$ and $\boldsymbol{E} = \{\mathbf{e}_{u, v} \in \mathbb{R}^q \mid (u, v) \in E\}$.
For \acrshort{gnn}s we often consider directed graphs, where the order of an edge $(u, v)$ defines $u$ as the sender and $v$ as the receiver.

A \acrshort{gnn} takes a labelled graph $\mathcal{G}$ and outputs a second graph $\mathcal{G}'$ with new labels.
Most \acrshort{gnn} architectures retain the same topology for $\mathcal{G}'$ as used in $\mathcal{G}$, in which case a \acrshort{gnn} can be viewed as a mapping from input labels $\mathcal{L}_\mathcal{G}$ to output labels $\mathcal{L}_\mathcal{G'}$.

A common \acrshort{gnn} framework is the \gls{mpnn} \citep{gilmer2017neural}, which generates this mapping using $\Tau$ steps or `layers' of computation.
At each layer $\tau \in \{0,\dots,\Tau-1\}$ in the network, a \textit{hidden state} $\mathbf{h}_v^{\tau+1}$ and \textit{message} $\mathbf{m}_v^{\tau+1}$ is computed for every node $v \in V$ in the graph.

An \gls{mpnn} implementation calculates these through its choice of \textit{message functions} and \textit{update functions}, denoted $M^\tau$ and $U^\tau$ respectively. A message function computes representations from hidden states and edge features, which are then aggregated and passed into an update function to compute new hidden states:
%
\begin{equation}
    \mathbf{m}_v^{\tau+1} =\!\!\! \sum_{u \in N(v)}\!\!M^\tau\left(\mathbf{h}_u^\tau, \mathbf{h}_v^\tau, \mathbf{e}_{u,v}\right),\quad \quad
    \mathbf{h}_v^{\tau+1} = U^\tau\left(\mathbf{h}_v^{\tau}, \mathbf{m}_v^{\tau+1}\right),
\end{equation}
%
for all nodes $v \in V$, where $N(v) = \{u \mid (u, v) \in E\}$ is the neighbourhood of all sender nodes connected to receiver $v$ by a directed edge.
The node input labels $\mathbf{v}_v$ are used as the initial hidden states $\mathbf{h}_v^0$. \gls{mpnn} assumes only \textit{node} output labels are required, using each final hidden state $\mathbf{h}_v^\Tau$ as the output label $\mathbf{v}'_v$.

\begin{figure}[b]
    \vskip -0.05in
    \centering
    \begin{subfigure}[t]{0.37\linewidth}
        \centering
        \includegraphics[width=0.9\linewidth]{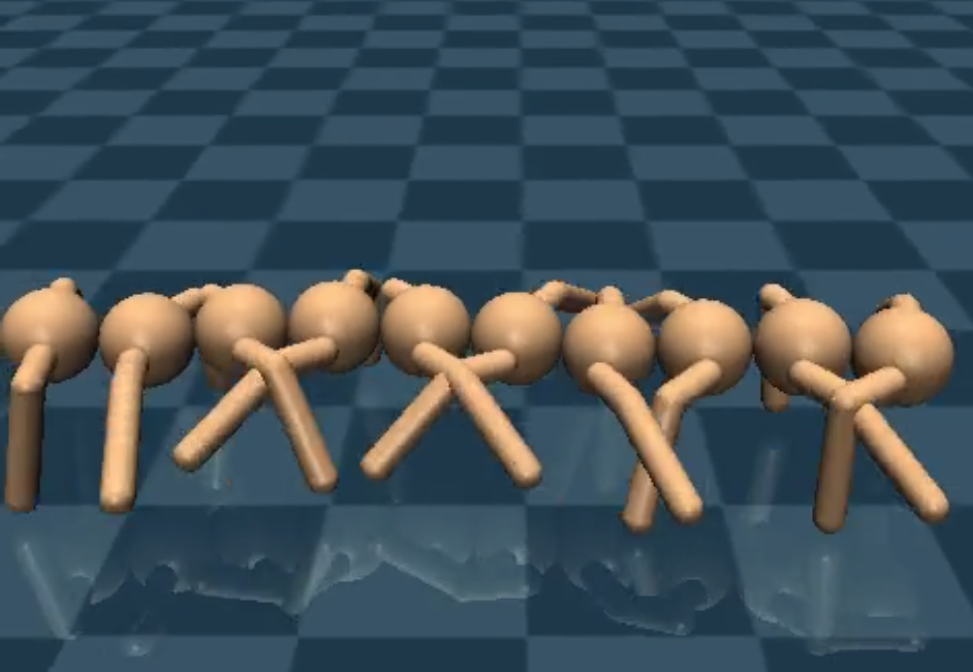}
    \end{subfigure}
    \hskip 0.3in
    \begin{subfigure}[t]{0.43\linewidth}
        \centering
        \includegraphics[width=0.9\linewidth]{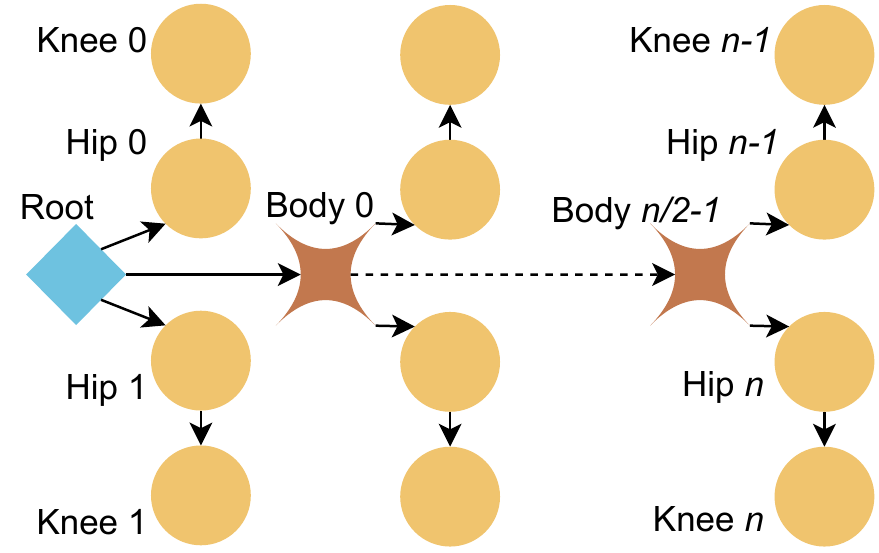}
    \end{subfigure}
    \vskip 0.05in
    \caption{A MuJoCo rendering of \texttt{Centipede-20} and its corresponding morphological graph.}
    \label{fig:centipede-n}
    \vskip -0.1in
\end{figure}

\subsection{NerveNet}
\label{NerveNet}

\acrshort{nn} is an \acrshort{mpnn} designed for locomotion control, based on the gated GNN  architecture \citep{li2015gated}. \acrshort{nn} uses the morphology (physical structure) of the agent as the basis for the \acrshort{gnn}'s input graph $\mathcal{G}$, with edges representing body parts and nodes representing the joints that connect them.

\cbb{
\acrshort{nn} assumes an MDP where the state $s$ can be factored into input labels $\boldsymbol{V}$, which are fed to the \acrshort{gnn} to generate output labels: $\mathbf{V}' = \textsc{NerveNet}(\mathcal{G}, \mathbf{V})$.
These are then used to parameterise a normal distribution defining the stochastic policy: $\pi(a\vert s)= {\mathcal {N}}(\mathbf{V}' , \text{diag}(\boldsymbol{\sigma}^{2}))$, where the standard deviation is a separate vector of parameters learned during training.
Actions $a$ are vectors, where each element represents the force to be applied at a given joint for the subsequent timestep.
The policy is trained using \acrshort{ppo}, with parameter updates computed via the Adam optimisation algorithm \citep{kingma2014adam}.
}

\cbb{
Internally, \acrshort{nn} uses an encoder $F_\text{in}$ to generate initial hidden states from input labels: $\mathbf{h}_v^0 = F_\text{in}\left(\mathbf{v}_v\right)$.
This is followed by a message function $M^\tau$ consisting of a single \acrshort{mlp} for all layers $\tau$ that takes as input only the state of the sender node: $\mathbf{m}_v^{\tau+1} = \sum_{u \in N(v)}\text{MLP}\left(\mathbf{h}_u^\tau\right)$.
The update function $U^\tau$ is a single \gls{gru} \citep{cho2014learning} that maintains an internal hidden state: $\mathbf{h}_v^{\tau+1} = \text{GRU}\left(\mathbf{m}_v^{\tau+1} \mid \mathbf{h}_v^{\tau}\right)$.
\acrshort{nn} propagates through $\Tau$ layers of message-passing and node-updating, before applying a decoder $F_\text{out}$ to turn final hidden states into scalar node output labels: $\mathbf{v}'_{v} = F_\text{out}(\mathbf{h}_v^T)$.
A diagram of the \acrshort{nn} architecture can be seen in Appendix \ref{Supplementary Figures}, Figure \ref{fig:mpnn-diagram-appdx}.
}




%% file: sections/Scaling.tex
In this section, we use \acrshort{nn} to analyse the challenges that limit \acrshort{gnn}s' ability to scale.
We focus on \acrshort{nn} as its architecture is more closely aligned with the \acrshort{gnn} framework than alternative approaches to structured locomotion control (see Section \ref{Related Work}).
We use mostly the same experimental setup as \citet{wang2018nervenet}, with details of any differences and our choice of hyperparameters outlined in Appendix \ref{Further Experimental Details}.


We focus on environments derived from the Gym \citep{brockman2016openai} suite, using the MuJoCo \citep{todorov2012mujoco} physics engine.
The main set of tasks we use to assess scaling is the selection of \texttt{Centipede-n} agents \citep{wang2018nervenet}, chosen because of their relatively complex structure and ability to be scaled up to high-dimensional input-action spaces.

The morphology of a \texttt{Centipede-n} agent consists of a line of \texttt{n/2} body segments, each with a left and right leg attached (see Figure \ref{fig:centipede-n}).
The graph used as the basis for the \gls{gnn} corresponds to the physical structure of the agent's body.
At each timestep in the environment, the MuJoCo engine sends a feature vector containing \cam{the positions of the agent's body parts and the forces acting on them}, expecting a vector to be returned specifying forces to be applied at each joint (full details of the state representation are given in Appendix \ref{Further Experimental Details}).
The agent is rewarded for forward movement along the $y$-axis as well as a small `survival' bonus for keeping its body within certain bounds, and given negative rewards proportional to the size of its actions and the magnitude of force it exerts on the ground.

\begin{figure}[b]
\vskip -0.1in
\begin{center}
\centerline{\includegraphics[width=\linewidth]{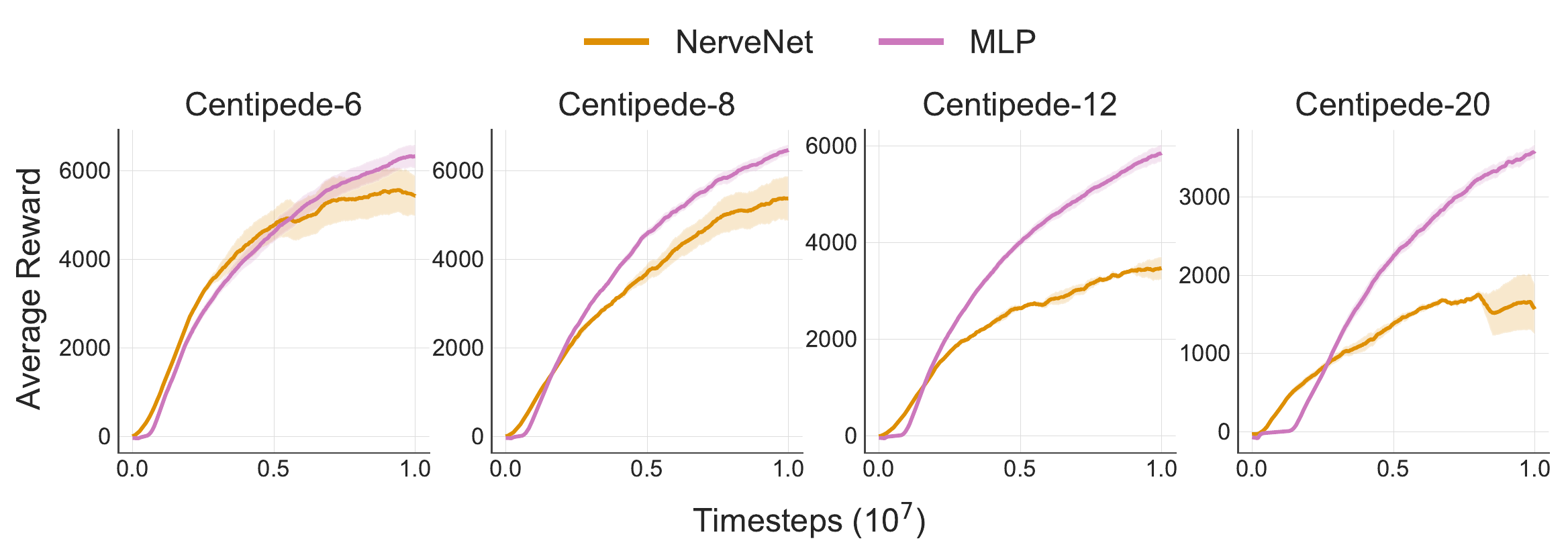}}
\caption{Comparison of the scaling of \acrshort{nn} relative to an \acrshort{mlp}-based policy. Performance is similar for the smaller agent sizes, but \acrshort{nn} scales poorly to the larger agents.}
\label{fig:baseline-scaling}
\end{center}
\vskip -0.25in
\end{figure}

Existing work applying \glspl{gnn} to locomotion control tasks avoid training directly on larger agents, i.e., those with many nodes in the underlying graph representation. For example, \citet{wang2018nervenet} state that for \acrshort{nn}, ``training a \texttt{CentipedeEight} from scratch is already very difficult''.
\citet{huang2020one} also limit training their \acrshort{smp} architecture to small agent types.

\subsection{Scaling Performance}

\cb{
To demonstrate the poor scaling of \acrshort{nn} to larger agents, we compare its performance on a selection of \texttt{Centipede-n} tasks to that of an \acrshort{mlp} policy.
Figure \ref{fig:baseline-scaling} shows that for the smaller \texttt{Centipede-n} agents both policies are similarly effective, but as the size of the agent increases, the performance of \acrshort{nn} drops relative to the \acrshort{mlp}.
A visual inspection of the behaviour of these agents shows that for \texttt{Centipede-20}, \acrshort{nn} barely makes forward progress at all, whereas the \acrshort{mlp} moves effectively.
}

As in previous literature \citep[e.g.,][]{wang2018nervenet,huang2020one}, we are ultimately not concerned with outperforming \acrshort{mlp}s on the specific training task, but rather matching their training task performance so that the \textit{additional} benefits of \acrshort{gnn}s can be realised.
\cam{In our setting we particularly wish to leverage the strong transfer benefits of \acrshort{gnn}s---as demonstrated by \citet{wang2018nervenet}---resulting from their capacity to process inputs of arbitrary size and structure.}

In other words, the focus of this paper is on deriving a method that can close the gap in Figure \ref{fig:baseline-scaling}, as doing so makes \acrshort{gnn}s a better choice overall given the trained policy transfers better than the \acrshort{mlp} equivalent \mi{(see Section \ref{sec:experiments} for experimental results)}.

\begin{table}[t]
  \vskip -0.1in
  \caption{KL-divergence from the policy before each update to the policy after, calculated over each batch. We train on $10^7$ timesteps, recording in the table the mean taken over last $10\%$ of steps.}
  \vskip 0.1 in
  \label{table:baseline-scaling-kl}
  \centering
  \begin{tabular}{lllll}
    \toprule
    & \multicolumn{4}{c}{Policy KL-divergence}              \\
    \cmidrule(r){2-5}
     Policy type & \texttt{Centipede-6} & \texttt{Centipede-8} & \texttt{Centipede-12} & \texttt{Centipede-20} \\
    \midrule
    \acrshort{mlp} & 0.021 & 0.024 & 0.031 & 0.044 \\
    \acrshort{nn} & 0.115 & 0.137 & 0.118 & 0.123 \\
    \bottomrule
  \end{tabular}
\end{table}

\subsection{Unstable policy updates}

\cb{
As outlined in Section \ref{ppo}, one of the key challenges for on-policy \acrshort{rl} is preventing individual updates from causing excessive changes in policy space (i.e., keeping it within the trust region).
Table \ref{table:baseline-scaling-kl} shows the extent to which this problem contributes to \acrshort{nn}'s poor scaling, calculating the average KL-divergence from the pre-update policy to the post-update policy for both policy types.
\acrshort{nn} has a consistently higher KL-divergence than the \acrshort{mlp} policy, indicating that \acrshort{ppo} finds it harder to ensure stable policy updates for the \acrshort{gnn}. 
}

\begin{figure}[b] 
\centering
\begin{minipage}{.4\linewidth}
  \centering
  \includegraphics[width=\linewidth]{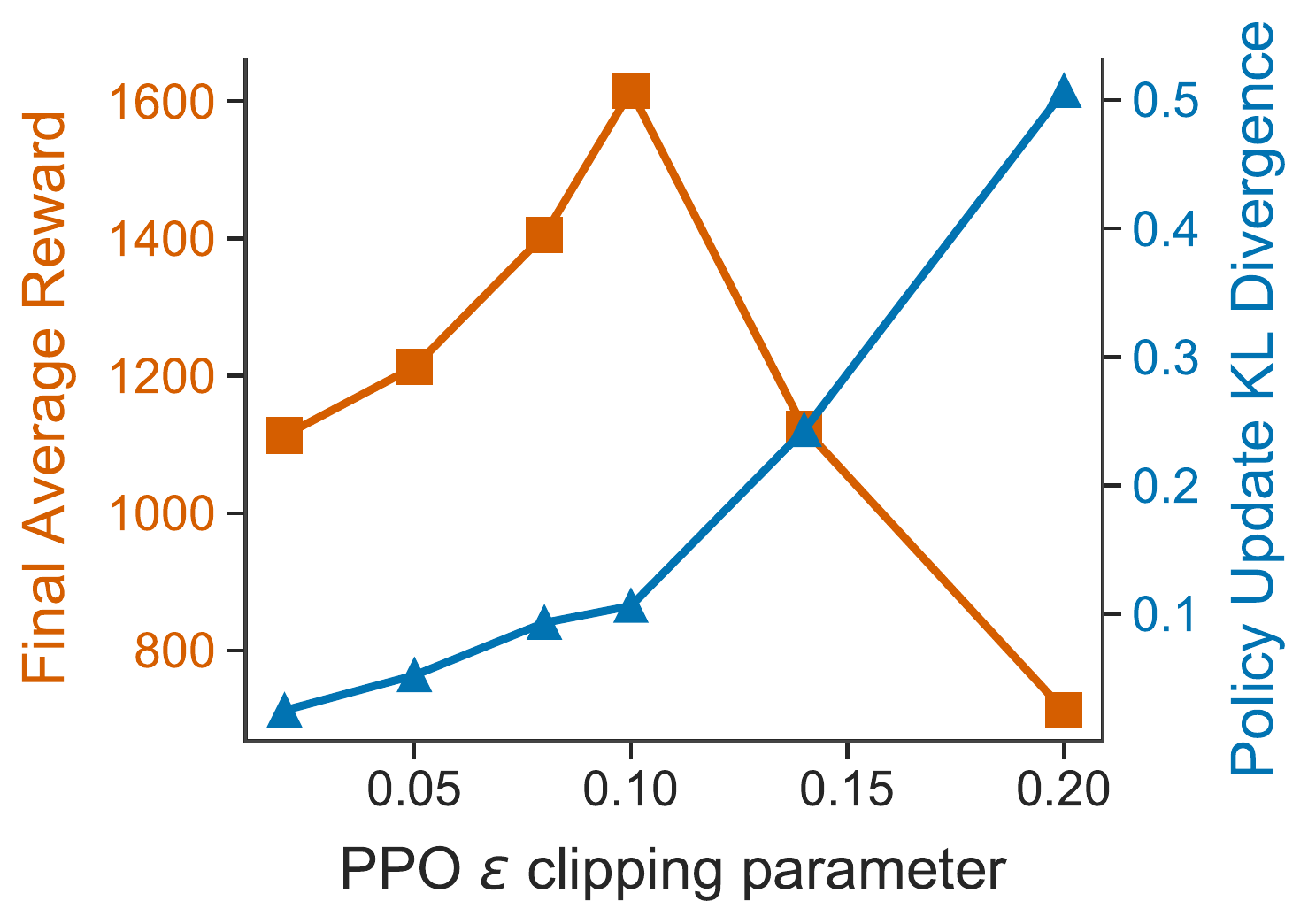}
  \captionof{figure}{Final performance of \acrshort{nn} on \texttt{Centipede-20} after ten million timesteps, across a range of $\epsilon$ clipping hyperparameter values. As $\epsilon$ increases (i.e., clipping is reduced) the KL divergence from the old to new policy (blue) increases. This improves performance (orange) up to a point, after which it begins to deteriorate.
  }
  \label{fig:kl-clipping}
\end{minipage}\qquad\begin{minipage}{.5\linewidth}
  \centering
  \includegraphics[width=\linewidth]{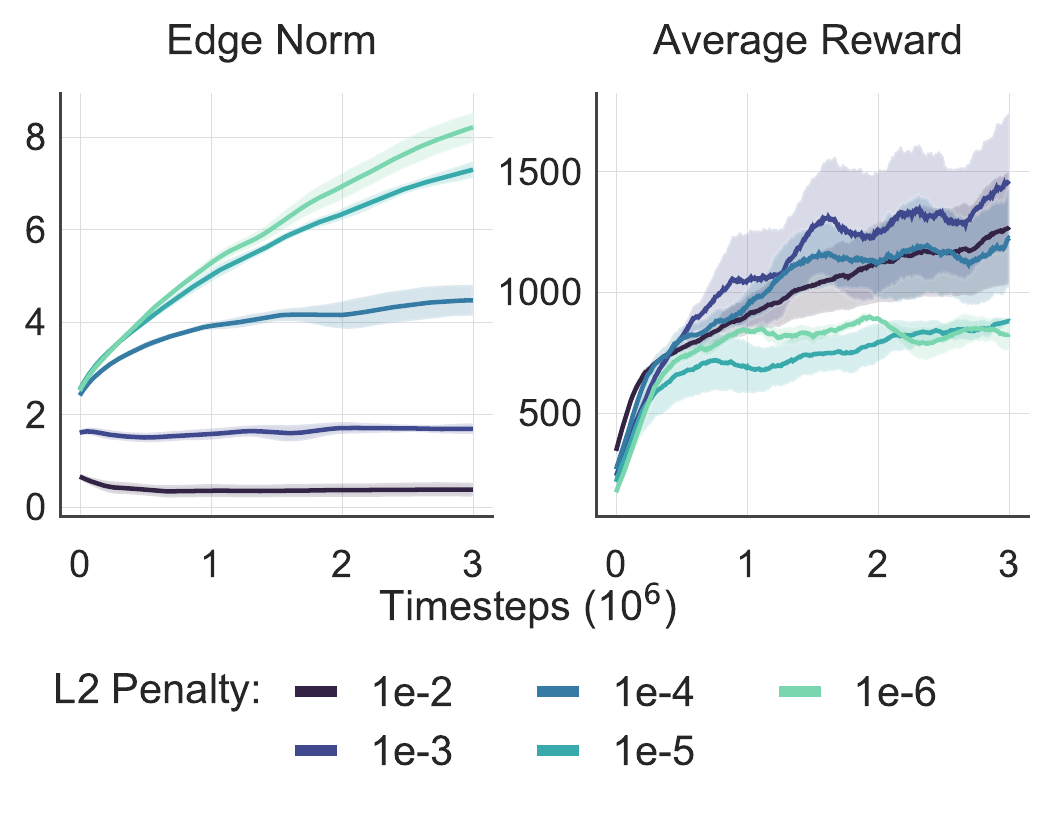}
  \vskip -0.1in
  \captionof{figure}{L2 regularisation for \acrshort{nn}'s message function across a range of values for the L2 penalty $\lambda$, trained on \texttt{Centipede-20}. Increasing this penalty reduces the L2 norm of the weights learned (left). Improved performance for higher values of $\lambda$ (right) indicates the presence of overfitting for the message function.}
  \label{fig:reg}
\end{minipage}
\end{figure}

\cbb{
We emphasise that this discrepancy persists even with carefully-tuned hyperparameter values for limiting policy divergence. Figure \ref{fig:kl-clipping} shows the performance of \acrshort{nn} across a range of PPO $\epsilon$-clipping values (see Section \ref{ppo}), and in all cases \acrshort{nn} is still substantially inferior to an \acrshort{mlp} (note that our experiments on \acrshort{nn} always use the best value of $\epsilon = 0.1$ found here).
As we demonstrate later (in Figure \ref{fig:buckets}), controlling policy divergence effectively is a key component in making \acrshort{gnn}s scale, but we see here that PPO alone does not control the divergence sufficiently to achieve this.
}

\subsection{Overfitting in \acrshort{nn}}

\begin{wrapfigure}{Rt}{0.45\linewidth}
    \vskip 0.16in
    \centering
    \includegraphics[width=\linewidth]{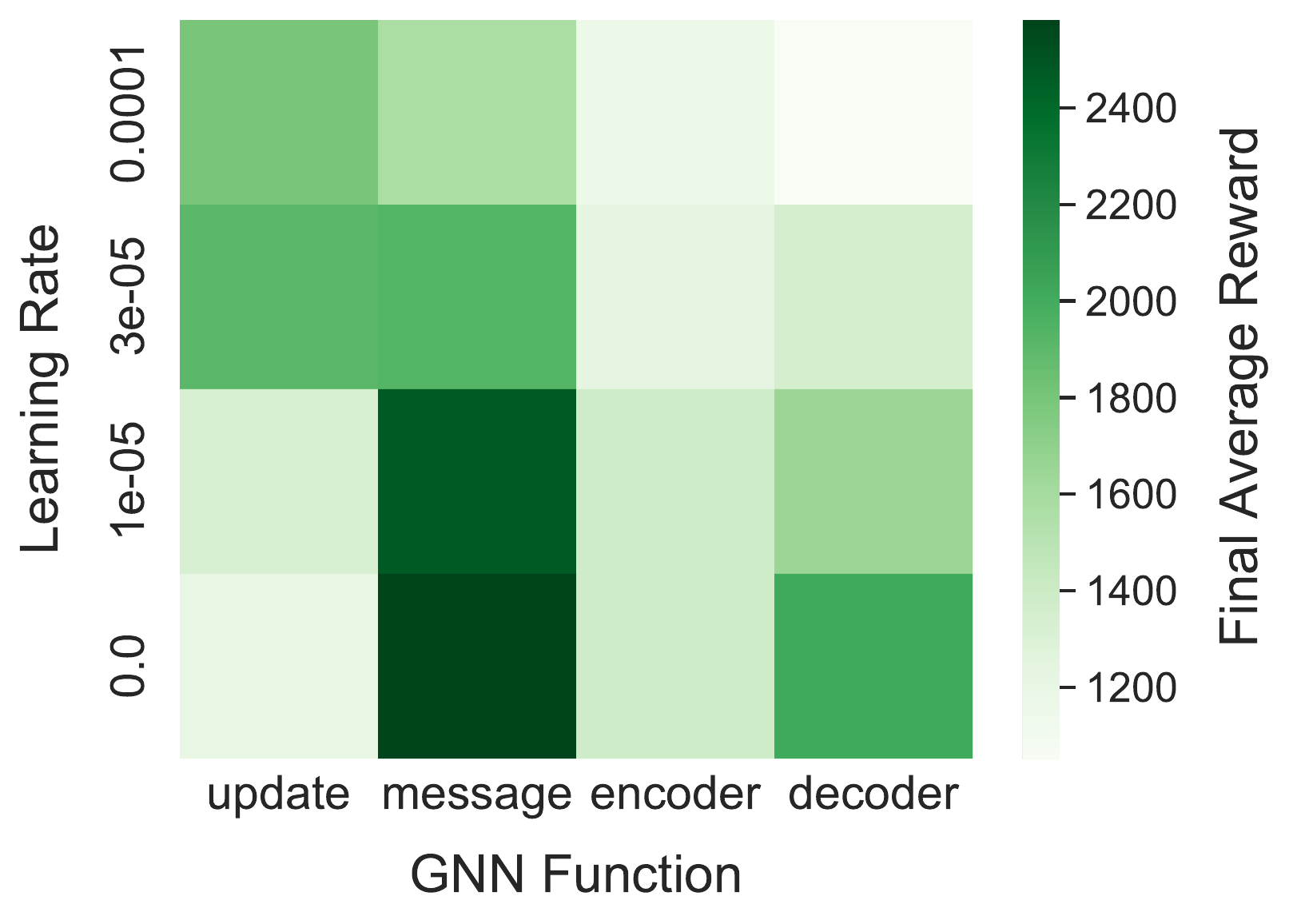}
    \caption{\mi{Colour-coded final \acrshort{nn} performance after 5M training steps on \texttt{Centipede-20} when changing learning rates for \textit{individual} \acrshort{gnn} components, compared to the base learning rate of $3 \times 10^{-4}$.}}
    \label{fig:learning_rates}
\end{wrapfigure}

\cb{
Excessive policy divergence resulting from updates can be understood as a form of overfitting.
\cb{Whereas the supervised interpretation of overfitting implies poor generalisation from training to test set, in this case we are concerned with poor generalisation across state-action distributions induced by different iterations of the policy during training.}
Specifically, each update involves an optimisation step aiming to increase the expected reward over a batch of trajectories generated using the \textit{pre-update} policy.
The challenge for \acrshort{rl} algorithms is that the agent is then evaluated and trained on trajectories generated using the \textit{post-update} policy, i.e., a different distribution to the one optimised on.
}

\cb{
For \acrshort{mpnn} architectures like \acrshort{nn}, it is a known deficiency that in the supervised setting, message functions implemented as \acrshort{mlp}s are prone to overfitting \citep[p.55]{hamilton2020graph}.
Here, we demonstrate that they also overfit (using the above interpretation) in our on-policy \acrshort{rl} setting.
Figure \ref{fig:reg} shows the effect of applying L2 regularisation (a standard approach to reducing overfitting) to the \acrshort{nn} architecture. We regularise the parameters $\boldsymbol{\theta}$ of \acrshort{nn}'s message function MLP $M_{\boldsymbol{\theta}}$, adding a $\lambda||\boldsymbol{\theta}||_2^2$ term to our objective function.
At the optimal value of $\lambda$ we see an improvement in performance (although still substantially inferior to using an \acrshort{mlp}), indicating that the unregularised message-passing \acrshort{mlp}s overfit. 
}

We also investigate lowering the learning rate in different parts of the \acrshort{gnn}, with the aim of identifying where overfitting is localised.
If parts of the network are particularly prone to damaging overfitting, training them more slowly may reduce their contribution to policy instability across updates.
Results for this experiment can be seen in Figure \ref{fig:learning_rates}.

Not only does lowering the learning rate in parts of the model improve performance, but surprisingly the best performance is obtained when the encoder $F_\text{in}$, message function $M$ and decoder $F_\text{out}$ each have their learning rate set to zero.
The encoder and decoder play a similar role to the message function, all of which are implemented as \acrshort{mlp}s, whereas the update function $U$ is a \acrshort{gru} \cb{(we experimented with using an \acrshort{mlp} update function, but found that this significantly reduced performance.)}.

\subsection{Snowflake}
\label{sec:sf}

Training with a learning rate of zero is equivalent to parameter freezing (e.g., \citet{DBLP:journals/corr/BrockLRW17}), where parameters are fixed to their initialised values throughout training.
\acrshort{nn} can learn a policy with some of its functions frozen, as learning still takes place in the un-frozen functions.
For instance, if we consider freezing the encoder, this results in an arbitrary mapping of input features to the initial hidden states.
As we still train the update function that processes this representation, so long as key information from the input features is not lost via the arbitrary encoding, the update function can still learn useful representations. The same logic applies to using a frozen decoder or message function.

Based on the effectiveness of parameter freezing within parts of the network, we propose a simple technique for improving the training of \acrshort{gnn}s via gradient-based optimisation, which we name \acrshort{sf} (a naturally-occurring frozen graph structure).
\cbb{\acrshort{sf} assumes a \acrshort{gnn} architecture made up internally of functions $F^1_{\boldsymbol{\theta}}, \dots, F^n_{\boldsymbol{\theta}}$, where $\boldsymbol{\theta}$ denotes the parameters of a given function.
Prior to training we select a fixed subset $\mathcal{Z} \subseteq \{F^1_{\boldsymbol{\theta}}, \dots, F^n_{\boldsymbol{\theta}}\}$ of these functions. Their parameters are then placed in  \acrshort{sf}'s \textit{frozen set} $\zeta = \{\boldsymbol{\theta} \mid F_{\boldsymbol{\theta}} \in Z\} $.}
During training, \acrshort{sf} excludes parameters in $\zeta$ from being updated by the optimiser, instead fixing them to whatever values the \acrshort{gnn} architecture uses as an initialisation.
Gradients still flow through these operations during backpropagation, but their parameters are not updated.
%
In practice, we found optimal performance for $\zeta = \{F_\text{in}, F_\text{out}, M^\tau\}$, i.e. when freezing the encoder, decoder and message function of the \gls{gnn}. If not stated otherwise, this is the architecture we refer to as \acrshort{sf} in subsequent sections.
A visual representation of \acrshort{sf} applied to the \acrshort{nn} model can be seen in Figure \ref{fig:sf-diagram-appdx}, Appendix \ref{Supplementary Figures}.

For our experiments, we initialise the values in the \acrshort{gnn} using the orthogonal initialisation \citep{saxe2014exact}.
We found this to be slightly more effective for frozen and unfrozen training than uniform and Xavier initialisations \citep{glorot2010understanding}.
For our message function, which has input and output dimensions of the same size, we find that performance with the frozen orthogonal initialisation is similar to that of simply using the identity function instead of an \acrshort{mlp}.
However, in the general case where the input and output dimensions of functions in the network differ (such as in the encoder and decoder, or in \acrshort{gnn} architectures where layers use representations of different dimensionality), this simplification is not possible and freezing is required.

%% file: sections/Related_Work.tex
\paragraph{Structured Locomotion Control}

Several different \acrlong{gnn}-like architectures \citep{scarselli2008graph, battaglia2018relational} have been proposed to learn policies for locomotion control.
\citet{wang2018nervenet} introduce \acrshort{nn}, which trains a \acrshort{gnn} based on the agent's morphology, along with a selection of scalable benchmarks.
\acrshort{nn} achieves multi-task and transfer learning across morphologies, even in the zero-shot setting (i.e., without further training), which standard \acrshort{mlp}-based policies fail to achieve.
\citet{DBLP:conf/icml/Sanchez-Gonzalez18} use a \acrshort{gnn}-based architecture for learning a model of the environment, which is then used for model-predictive control.

\citet{huang2020one} propose \gls{smp}, which focuses on multi-task training and shows strong generalisation to out-of-distribution agent morphologies using a single policy.
The architecture of \gls{smp} has similarities with a \acrshort{gnn}, but requires a tree-based description of the agent's morphology, and replaces size- and permutation-invariant aggregation with a fixed-cardinality \acrshort{mlp}.
\citet{DBLP:conf/nips/PathakLDIE19} propose \gls{dgn}, where a \acrshort{gnn} is used to learn a policy enabling multiple small agents to cooperate by combining their physical structures.

Amorpheus
\citep{kurin2020my} uses an architecture based on transformers \citep{NIPS2017_3f5ee243} to represent locomotion policies.
Transformers can be seen as \acrshort{gnn}s using attention for edge-to-vertex aggregation and operating on a fully connected graph,
meaning computational complexity scales quadratically with the graph size.

For all of these existing approaches to \acrshort{gnn}-based locomotion control, training is restricted to small agents. In the case of \acrshort{nn} and \gls{dgn}, emphasis is placed on the ability to perform zero-shot transfer to larger agents, but this still incurs a significant drop in performance.

\paragraph{Graph-Based Reinforcement Learning}

\glspl{gnn} have recently gained traction in~\gls{rl} due to their support for variable sized inputs and outputs, enabling new \gls{rl} applications and enhancing the capabilities of agents on existing benchmarks.

\citet{DBLP:conf/nips/KhalilDZDS17} apply DQN~\cite{dqn} to combinatorial optimisation problems using Structure2Vec~\cite{structure2vec} for function approximation.
\citet{gil2020} use policy gradient methods to learn heuristics of a quantified Boolean formulae solver, while \citet{DBLP:conf/nips/KurinGWC20} use DQN~\cite{dqn} with graph networks~\citep{battaglia2018relational} to learn the branching heuristic of a Boolean SAT solver.
\cam{\citet{DBLP:conf/nips/KlissarovP20} use a \acrshort{gnn} to represent an MDP, which is then used to learn a form of reward shaping. \citet{deac2020xlvin} similarly use a \acrshort{gnn} MPD representation to generalise Value Iteration Nets \citep{DBLP:conf/nips/TamarLAWT16} to a broader class of MDPs.}

Other approaches involve the construction of graphs based on factorisation of the environmental state into objects with associated attributes~\citep{bapst2019structured, loynd2020working}.
In multi-agent~\gls{rl}, researchers have used a similar approach to model the relationship between agents, as well as between environmental objects~\citep{DBLP:conf/iclr/ZambaldiRSBLBTR19, DBLP:conf/icml/IqbalWPBWS21, li2020deep}.
In this setting, increasing the number of agents can result in additional challenges, such as combinatorial explosion of the action space. Our approach can be potentially useful to the above work, in improving scaling properties across a variety of domains.

\paragraph{Random Embeddings and Parameter Freezing}

\cb{
Embeddings represent feature vectors that have been projected into a new, typically lower-dimensional space that is easier for models to process.
\citet{DBLP:conf/kdd/BinghamM01} show that multiplying features by a randomly generated matrix (e.g., with entries sampled from a Gaussian distribution) preserves similarity well and empirically attains comparable performance to PCA.
\citet{DBLP:journals/jair/WangHZMF16} use this approach to apply Bayesian Optimisation to high dimensional datasets by randomly projecting them into a smaller subspace.
For natural language applications, commonly used pre-trained embeddings (e.g., word2vec \citep{DBLP:conf/emnlp/PenningtonSM14}, GloVe \citep{DBLP:journals/corr/abs-1301-3781}) have been shown to offer only a small benefit over random embeddings on benchmark datasets \citep{DBLP:conf/icon-nlp/KocmiB17, DBLP:journals/corr/DhingraLSC17} and may offer no benefit on industry-scale data \citep{DBLP:conf/acl/AroraMZR20}.
}

\cb{
More generally, random embeddings can be induced by freezing typically-learned parameters within a model to fixed values throughout training.
This approach has been explored for transformer architectures, where fixed attention weights (either Gaussian-distributed \citep{DBLP:conf/acl/YouSI20} or hand-crafted \citep{DBLP:conf/emnlp/RaganatoST20}) show no significant drop in performance, and even freezing intermediate feedforward layers still enables surprisingly effective learning \citep{DBLP:journals/corr/abs-2103-05247}.
A similar technique can also be found in common fine-tuning methods, where parameters are pre-trained on another, possibly unsupervised objective, but frozen during training except for the final layer
\citep[e.g.,][]{DBLP:conf/nips/YosinskiCBL14, DBLP:conf/icml/HoulsbyGJMLGAG19}.
}

%% file: sections/Experiments.tex
\label{sec:experiments}


\begin{figure*}[b]
\vskip -0.15in
\begin{center}
\centerline{\includegraphics[width=\linewidth]{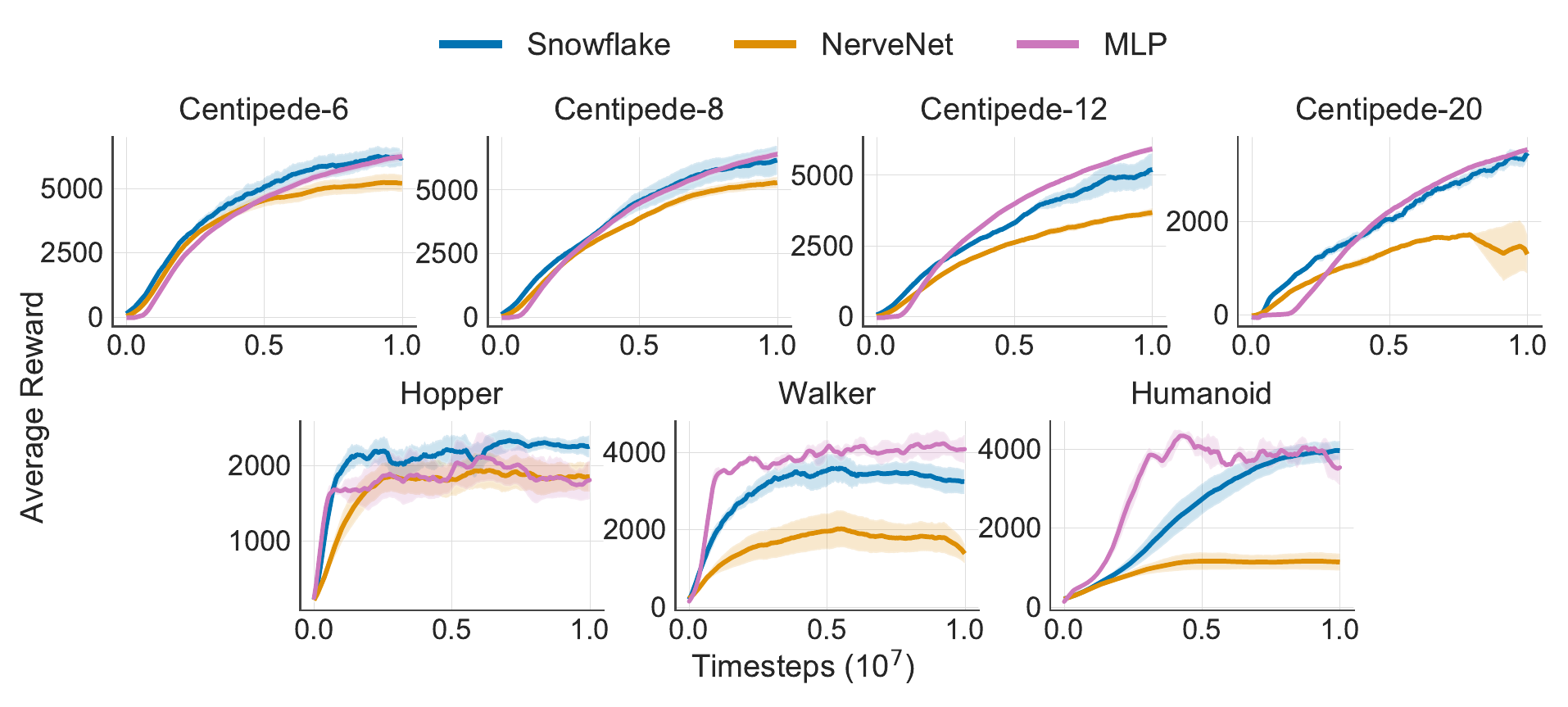}}
\caption{Comparison of the performance of \acrshort{sf} training, regular \acrshort{nn} and the \acrshort{mlp}-based policy. \acrshort{sf} enables effective scaling to the larger agents, significantly outperforming regular \acrshort{nn} and comparable to using an \acrshort{mlp}-based policy.}
\label{fig:scaling-tasks}
\end{center}
\vskip -0.25in
\end{figure*}

We present experiments evaluating the performance of \acrshort{sf} when applied to \acrshort{nn}, and compare against regular \acrshort{nn} and \acrshort{mlp} policies.
We evaluate each model on a selection of MuJoCo tasks, including three standard tasks from the Gym suite \citep{brockman2016openai} and the \texttt{Centipede-n} agents from \citet{wang2018nervenet}.
\cbb{Note that we do not train on even larger \texttt{Centipede-n} agents due to wall-clock simulation time becoming prohibitively large.}


All training statistics are calculated as the mean across six independent runs (unless specified otherwise), with the standard error across runs indicated by the shaded areas on each graph.
The average reward typically has high variance, so to smooth our results we plot the mean taken over a sliding window of 30 data points.
Further experimental details are outlined in Appendix \ref{Further Experimental Details}.

\paragraph{Scaling to High-Dimensional Tasks}
Figure~\ref{fig:scaling-tasks} compares the scaling properties of the regular \acrshort{nn} model with \acrshort{sf}.
As the size of the agent increases, \acrshort{sf} significantly outperforms \acrshort{nn} with comparable asymptotic performance to the \acrshort{mlp}.
\cb{This indicates that \acrshort{sf} is successful in addressing the deficiencies of regular \acrshort{nn} training, and that freezing overfitting parameters is an effective training strategy in this setting.
This holds true across locomotive agents with substantially different morphologies.}

\paragraph{Zero-shot transfer}

\begin{figure*}[t]
\centering
\begin{minipage}{.43\linewidth}
  \centering
  \includegraphics[width=\linewidth]{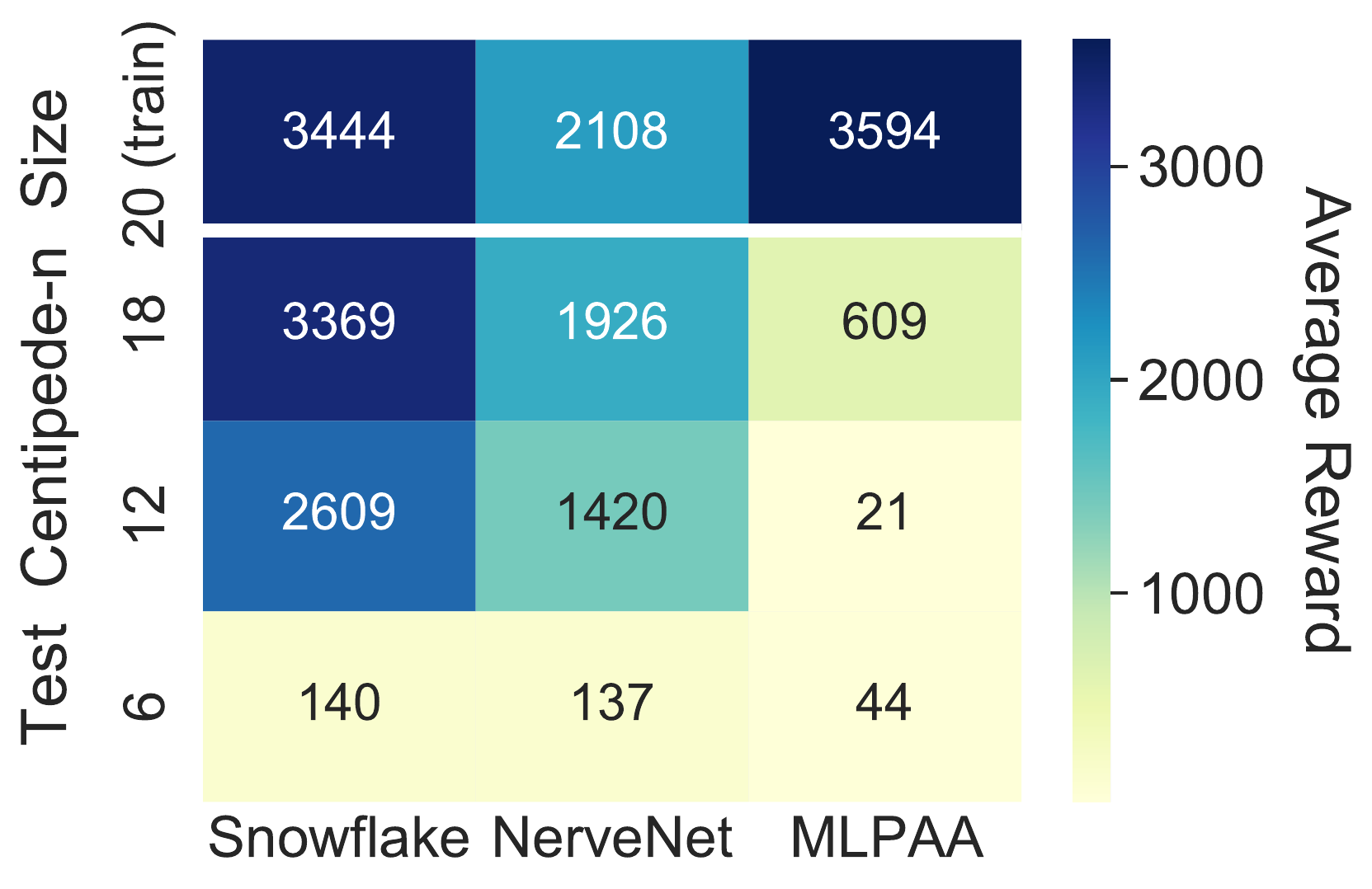}
  \captionof{figure}{Zero-shot transfer performance for \acrshort{sf}, \acrshort{nn}, and \acrshort{mlp} models trained on \texttt{Centipede-20}, evaluated across a range of sizes.
  }
  \label{fig:transfer}
\end{minipage}\quad\begin{minipage}{.53\linewidth}
  \centering
  \includegraphics[width=\linewidth]{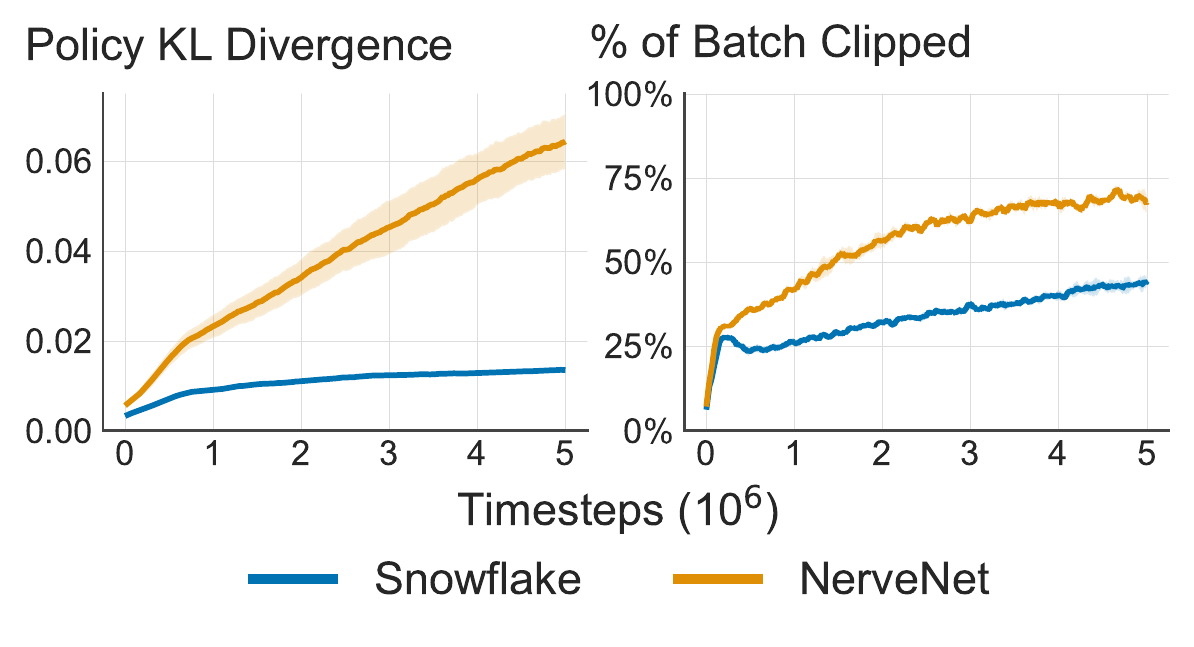}
  \captionof{figure}{The effect of \acrshort{sf} on policy divergence and \acrshort{ppo} clipping on \texttt{Centipede-20}. By freezing parts of the network that overfit, \acrshort{sf} reduces the policy KL divergence leading to less clipping during training.}
  \label{fig:buckets}
\end{minipage}
    \vskip 0.15in
\end{figure*}

\cb{
An important motivation for improving \acrshort{gnn} scaling
is to harness their transfer capabilities on large tasks.
Regular \acrshort{nn} is limited by the fact that it can only effectively train on and transfer between small agent sizes.\footnote{We found that a regular \acrshort{nn} policy trained to achieve high training task performance on a smaller agent (e.g., \texttt{Centipede-6}) does not transfer effectively to larger ones, as reflected in \citep[Figure 4]{wang2018nervenet}.}
We show in Figure \ref{fig:transfer} that \acrshort{sf} attains exceptional zero-shot transfer performance across centipede sizes, surpassing alternative methods.
\acrshort{sf} is the only method that can train a single policy that is effective on \texttt{Centipede-20} through to \texttt{12}.
}

\cb{
 \acrshort{sf}  therefore achieves our initial objective: combining the strong training task performance of an \acrshort{mlp} and the strong transfer performance of regular \acrshort{nn}.
As a consequence, \acrshort{sf}-trained \acrshort{gnn}s offer the most promising policy representation for locomotion control tasks where transfer is a desirable property.
}

\paragraph{Policy Stability and Sample Efficiency}

\begin{figure}[t]
    \centering
    \includegraphics[width=0.63\linewidth]{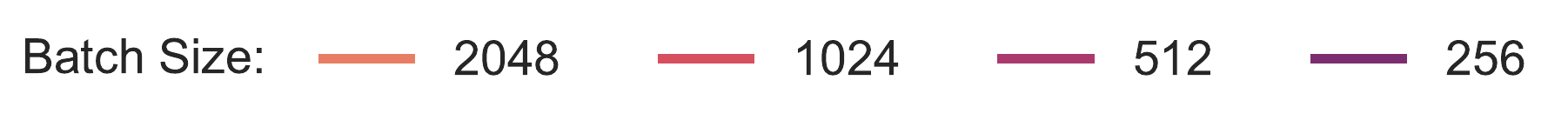}
    \begin{subfigure}[t]{0.507\linewidth}
        \centering
        \includegraphics[width=\linewidth]{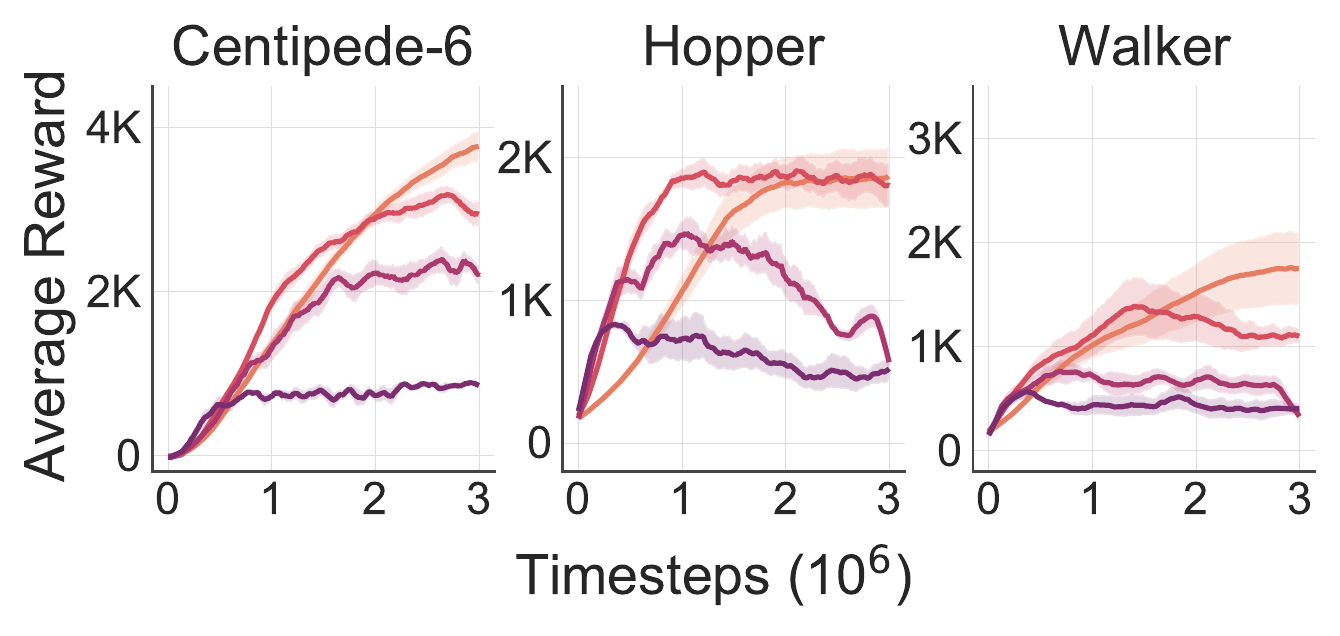}
        \caption{\acrshort{nn}}
    \end{subfigure}%
    \hskip 0.015in
    \vrule
    \hskip 0.015in
    \begin{subfigure}[t]{0.473\linewidth}
        \centering
        \includegraphics[trim={0.8cm 0 0 0}, clip, width=\linewidth]{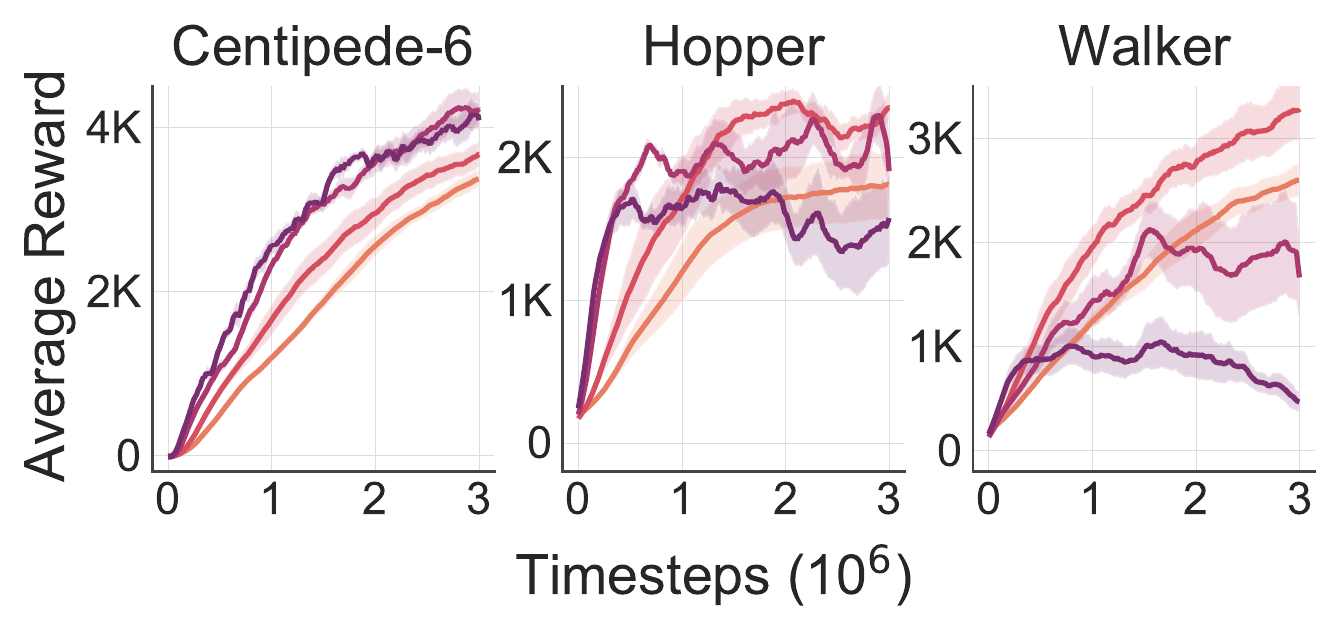}
        \caption{\acrshort{sf}}
    \end{subfigure}
    \caption{Effectiveness of \acrshort{sf} across smaller batch sizes relative to standard \acrshort{nn} training. \acrshort{sf} is able to use smaller batch sizes, leading to improved sample efficiency. This is due to \acrshort{sf} reducing policy divergence across updates. Corresponding policy divergence plots can be found in the appendix.}
    \label{fig:se-tasks}
\end{figure}

By reducing overfitting in parts of the \acrshort{gnn}, \acrshort{sf} mitigates the effect of harmful policy updates seen with regular \acrshort{nn}. As a consequence, the policy can train effectively on smaller batch sizes.
This is demonstrated in Figure \ref{fig:se-tasks}, which shows the performance of \acrshort{nn} trained regularly versus using \acrshort{sf} as the batch size decreases.

A potential benefit of training with smaller batch sizes is improved sample efficiency, as fewer timesteps are taken in the environment per update.
However, smaller batch sizes also lead to increased policy divergence
\cb{due to increased noise in the gradient estimate.}
When the policy divergence is too great, performance begins to decrease, limiting how small the batch can be.
\cb{However due to a reduction in policy divergence as a result of \acrshort{sf}, we can afford to use smaller batch sizes while still keeping the policy under control.}
This provides a wider motivation for the use of \acrshort{sf} than just scaling to larger agents: it also improves sample efficiency across agents regardless of size.

The success of \acrshort{sf} in scaling to larger agents can also be understood in this context.
Without \acrshort{sf}, for \acrshort{nn} to attain strong performance on large agents an infeasibly large batch size would be required, leading to poor sample efficiency.
The more stable policy updates enabled by \acrshort{sf} make solving these large tasks tractable.

\paragraph{PPO Clipping}

\acrshort{sf}'s improved policy stability also reduces the amount of clipping performed by \acrshort{ppo}
across each training batch.
Figure \ref{fig:buckets} shows the percentage of state-action pairs that are clipped for regular \acrshort{nn} versus \acrshort{sf} on the \texttt{Centipede-20} agent, as a result of reduced KL divergence\footnote{\cam{It may seem counter-intuitive that the KL divergence increases over time. This is due to the standard deviation of the policy (a learned parameter) reducing as training progresses, with the agent trading exploration for exploitation.}}.

When \acrshort{nn} is trained without using \acrshort{sf} a larger percentage of state-action pairs are clipped during \acrshort{ppo} updates---a consequence of the greater policy divergence caused by overfitting.
For \acrshort{ppo} if too many data points reach the clipping limit during optimisation, the algorithm is only able to learn on a small fraction of the experience collected, reducing the effectiveness of training.
One of \acrshort{sf}'s strengths is that because it reduces policy divergence it requires less severe restrictions to keep the policy within the trust region.
The combination of this effect and the ability to train well on smaller batch sizes enables \acrshort{sf}'s strong performance on the largest agents.

%% file: sections/Conclusion.tex
We proposed \acrshort{sf}, a method that enables \acrshort{gnn}-based policies to be trained effectively on much larger locomotive agents than was previously possible.
We no longer observe a substantial difference in performance between using \acrshort{gnn}s to represent a locomotion policy and the standard approach of using \acrshort{mlp}s, even on the most challenging morphologies.
As a consequence, \acrshort{gnn}s now offer an alternative to \acrshort{mlp}s for more than just simple tasks, and if the additional features of \acrshort{gnn}s such as strong transfer are a requirement, then they are likely to be a more effective choice.
We have also provided insight into why poor scaling occurs for certain \acrshort{gnn} architectures, and why parameter freezing is effective in addressing the overfitting problem we identify.

\cam{Limitations of our work include the upper-limit on the size of agents we were able to simulate, the use of a single algorithm and architecture, and a focus only on locomotion control tasks.
Future work may include applying alternative RL algorithms and GNN architectures, schemes for automating the selection of frozen parts of the network, and applying \acrshort{sf}-like methods to a wider range of learning problems.}

%% file: sections/Appendix.tex
\subsection{Ethical Discussion}
\label{Ethics}

Our work addresses the problem of scaling GNNs for simulated locomotion control. As our data is generated and models trained solely in a simulated physics engine, the direct ethical implications of our work are minimal. However, we identify a number of potential risks emerging from extensions and alternative applications of our work. These centre on safety and bias concerns relating to robotic control, and transferring trained policies to new tasks/agents.

\paragraph{Robotic Control} As we only ever conduct rollouts of the policy in simulation, our agent is not trained with any safety constraints in mind, which would likely be a requirement for real-world applications. Safety is particularly relevant in the wider context of our work, as the aim of scaling GNNs to more complex and capable agents potentially gives rise to increasingly unsafe behaviours and outcomes in the worst-case. Future work is required to assess if existing RL safety methods \citep[see][]{DBLP:journals/jmlr/GarciaF15} are as effective when GNN policies are used.

Although there are many beneficial use-cases for robotic agents, there is also potential for negative social outcomes. These may be through agents that are designed directly to do harm such as autonomous weapons, or that are used in socially irresponsible applications. We encourage researchers who use our methods in the pursuit of enabling new robotics applications to give consideration to such outcomes.

\paragraph{Policy Transfer} Although effective transfer has the benefit of reducing the need for further training on the target task, the resulting policy is inevitably biased towards the original training task. Algorithmic bias has been highlighted as a key challenge in recent years for AI fairness, particularly in the supervised setting \citep{DBLP:journals/corr/abs-1908-09635}, but also for RL algorithms \citep{DBLP:conf/icml/JabbariJKMR17}.

In the case of GNN policy representations, this problem can arise if the graphs (and associated labels) trained on contain harmful bias. For instance, consider a GNN policy trained on (graph-based) traffic data to optimise an RL objective, such as cumulative journey time for route planning. If the training data consists only of roads in certain geographical areas, then transferring the policy to out-of-distribution areas may lead to unsuitable actions and a disparity in outcomes. Safety constraints satisfied on the training roads (e.g. limiting the number of road accidents) may also no longer be satisfied when transferring to new areas.

\paragraph{Ethical Conduct} Our training data consists entirely of simulated physical observations from the MuJoCo environment. There is no human-generated data used in our research, nor does any of our data relate to real-world phenomena (beyond the laws of physics and design of our agents). We are therefore satisfied that our use of data is appropriate and ethical.

\subsection{Further Experimental Details}
\label{Further Experimental Details}

Here we outline further details of our experimental approach to supplement those given in Section \ref{sec:experiments}.

\subsubsection*{Data Generation}

As typical when training PPO on simulated environments, we train a policy by interleaving two processes:
first, we perform repeated rollouts of the current policy in the environment to generate on-policy training data, and second, we optimise the policy with respect to the training data collected to generate a new policy, then repeat.

To improve wall-clock training time, for larger agents we perform rollouts in parallel over multiple CPU threads, scaling from a single thread for \texttt{Centipede-6} to five threads for \texttt{Centipede-20}.
Rollouts terminate once the sum of timesteps experienced across all threads reaches the training batch size.
\cam{For our experiments the main computational cost as the agent size scales is the simulator, not the training of the network. Our GNN implementation is therefore not highly optimised as this is not our bottleneck.}

For optimisation we shuffle the training data randomly and split the batch into eight minibatches.
We perform ten optimisation epochs over these minibatches, in the manner defined by the \acrshort{ppo} algorithm \citep{schulman2017proximal} (see Section \ref{ppo}).

Each experiment is performed six times and results are averaged across runs. The exceptions to this are Figure \ref{fig:learning_rates} where results are an average of three runs, and the \texttt{Centipede-n} tasks in Figure \ref{fig:scaling-tasks} where results are an average of ten runs.

\subsubsection*{Hyperparameter Search}

Our starting point for selecting hyperparameters is the hyperparameter search performed by \citet{wang2018nervenet}, whose codebase ours is derived from.

To ensure that we have the best set of hyperparameters for training on large agents, we ran our own hyperparameter search on \texttt{Centipede-20} for \acrshort{sf}, as seen in Table \ref{table:hyp}.

\begin{table}[h]
\vskip 0.15in
\begin{center}
 \begin{tabular}{c | c} 
 \hline
 Hyperparameter & Values \\
 \hline
 Batch size & 512, 1024, \textbf{2048}, 4096 \\ 
 Learning rate & 1e-4, \textbf{3e-4}, 1e-5 \\
 Learning rate scheduler & adaptive, \textbf{constant} \\
 $\epsilon$ clipping & 0.02, 0.05, \textbf{0.1}, 0.2 \\
 \acrshort{gnn} layers & 2, \textbf{4}, 10 \\
 \acrshort{gru} hidden state size & \textbf{64}, 128 \\
 Learned action std & shared, \textbf{separate} \\
 \hline
\end{tabular}
\end{center}
\caption{Hyperparameter search for \acrshort{sf} on \texttt{Centipede-20}. Values in bold resulted in the best performance.}
\label{table:hyp}
\end{table}

Across the range of agents tested on, we conducted a secondary search over just the batch size, learning rate and $\epsilon$ clipping value for each model. For the latter two hyperparameters, we found that the values in Table \ref{table:hyp} did not require adjusting.

For the batch size, we used the lowest value possible until training deteriorated. Using \acrshort{nn}, a batch size of 2048 was required throughout, whereas using \acrshort{sf} a batch size of 1024 was best for \texttt{Walker},  \texttt{Centipede-20} and \texttt{Centipede-12}, 512 was best for \texttt{Centipede-8} and \texttt{Centipede-6}, and 2048 for all other agents.

\citet{wang2018nervenet} provide experimental results for the \acrshort{nn} model, which we use as a baseline for our experiments.
Out of the \texttt{Centipede-n} models, they provide direct training results for \texttt{Centipede-8} (see the non-pre-trained agents in their Figure 5). Our performance results are comparable, but taken over many more timesteps. Their final \acrshort{mlp} results appear slightly different to ours at the same point (they attain roughly 500 more reward), likely due to hyperparameter tuning for performance over a different time-frame.

They also provide performance metrics for trained \texttt{Centipede-4} and \texttt{Centipede-6} agents across the models compared (their Table 1).
The results reported here are significantly less than the best performance we attain for both \acrshort{mlp} and \acrshort{nn} on \texttt{Centipede-6}.
We suspect this discrepancy is due to running for fewer timesteps in their case, but precise stopping criteria is not provided.

\subsubsection*{Computing Infrastructure}

Our experiments were run on four different machines during the project, depending on availability.
These machines use variants of the Intel Xeon E5 processor (models 2630, 2699 and 2680), containing between 44 and 88 CPU cores.
As running the agent in the MuJoCo environment is CPU-intensive, we observed little decrease in training time when using a GPU; hence the experiments reported here are only run on CPUs.

Runtimes for our results vary significantly depending on the number of threads allocated and batch size used. Our standard runtime for \texttt{Centipede-6} (single thread) for ten million timesteps is around 24 hours, scaling up to 48 hours for our standard \texttt{Centipede-20} configuration (five threads).
Our experiments on the default MuJoCo agents also take approximately 24 hours for a single thread.

\subsubsection*{State Space Description}

\cam{The following is a breakdown of the information sent by the environment at each timestep to the different MuJoCo node types for the \texttt{Centipede-n} benchmark. Each different \texttt{body} and \texttt{joint} node receives its own version of this set of data:}

\begin{table}[h]
\vskip 0.15in
\begin{center}
 \begin{tabular}{c | c | c} 
 \hline
 Node Type & Observation Type & Axis \\
 \hline
 \hline
 \multirow{6}{*}{body} & force & x\\
 & force & y\\
 & force & z\\
 & torque & x\\
 & torque & y\\
 & torque & z\\
 \hline
 \multirow{2}{*}{joint} & position & x\\
 & velocity & x\\
 \hline
 \multirow{17}{*}{root} & orientation & x\\
 & orientation & y\\
 & orientation & z\\
 & orientation & a\\
 & velocity & x\\
 & velocity & y\\
 & velocity & z\\
 & angular velocity & x\\
 & angular velocity & y\\
 & angular velocity & z\\
 & position & z\\
 & force & x\\
 & force & y\\
 & force & z\\
 & torque & x\\
 & torque & y\\
 & torque & z\\
 \hline
\end{tabular}
\end{center}
\caption{Description of the state space.}
\label{table:env-state}
\end{table}

\cam{The root's z-position (height) is relative to the (global) floor of the environment. For this benchmark the joints are hinge joints, meaning that there is only one degree of freedom, and its position value reflects the joint angle (note that x-axis here refers to the joint's relative axis, not the global coordinate frame).}

\cam{Our algorithm only strictly considers observations to come from joints rather than from body and root nodes. In this we follow the example set by \acrshort{nn}, which for the sake of simplicity concatenates body node observations with neighbouring joint observations, treating the resulting vector as a combined joint representation, which is then fed to the GNN.}

\subsection{Sources}
\label{Sources}

\cam{Our source code can be found at \url{https://github.com/thecharlieblake/snowflake/}, alongside documentation for building the software and its dependencies. Our code is an extension of the \acrshort{nn} codebase: \url{https://github.com/WilsonWangTHU/NerveNet}.
This repository contains the original code/schema defining the \texttt{Centipede-n} agents.}

The other standard agents are taken from the Gym \citep{brockman2016openai}: \url{https://github.com/openai/gym}. The specific hopper, walker and humanoid versions used are \texttt{Hopper-v2}, \texttt{Walker2d-v2} and \texttt{Humanoid-v2}.

For our \acrshort{mlp} results on the Gym agents, as state-of-the-art performance baselines have been well established in this case, we use the OpenAi Baselines codebase (\url{https://github.com/openai/baselines}) to generate results, to ensure the most rigorous and fair comparison possible.

The MuJoCo \citep{todorov2012mujoco} simulator can be found at: \url{http://www.mujoco.org/}.
Note that a paid license is required to use MuJoCo.
The use of free alternatives was not viable in our case as our key benchmarks are all defined for MuJoCo.

\newpage
\subsection{Supplementary Figures}
\label{Supplementary Figures}

\begin{figure}[h]
\begin{center}
\vskip 0.25in
\centerline{\includegraphics[width=0.92\linewidth]{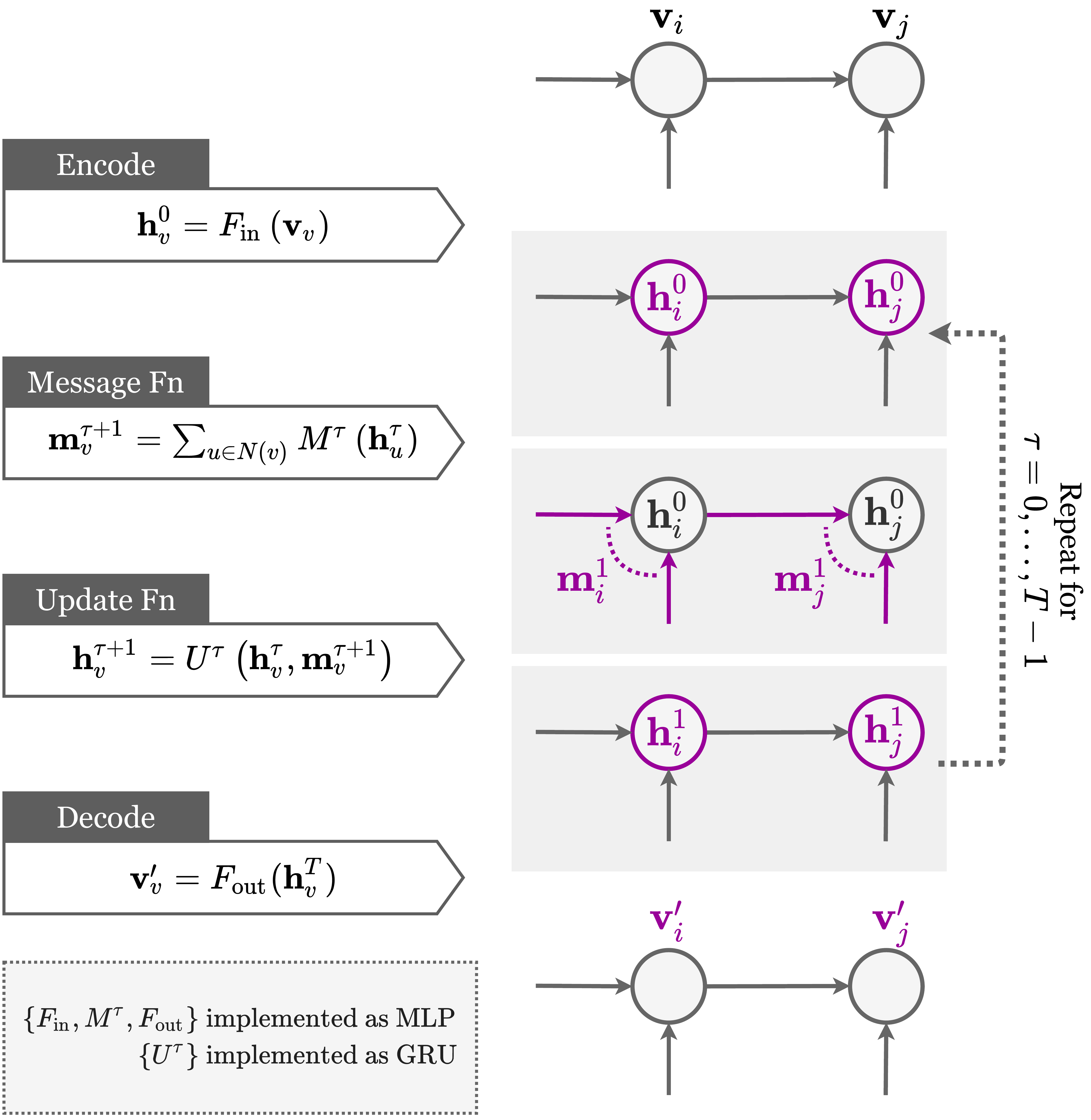}}
\vskip 0.45in
\caption{A visual representation of the \acrshort{nn} architecture. Updated representations at each step are indicated in purple. Given an input vector $\mathbf{v}$ at each node, \acrshort{nn} computes scalar outputs $\mathbf{v}'$ through a series of propagation steps. Initially, the encoder is used to compute hidden states $\mathbf{h}$ at each node. These are passed into the message function, which computes an incoming message $\mathbf{m}$ for each node based on the hidden states of its neighbours. The update function then computes a new hidden state representation for each node based on the incoming message and the previous hidden state. The message function and update function then repeat their operations $T$ times, before feeding the final hidden states to the decoder, which produces outputs $\mathbf{v}'$.}
\label{fig:mpnn-diagram-appdx}
\end{center}
\end{figure}

\begin{figure}[h]
\begin{center}
\vskip 0.25in
\centerline{\includegraphics[width=0.75\linewidth]{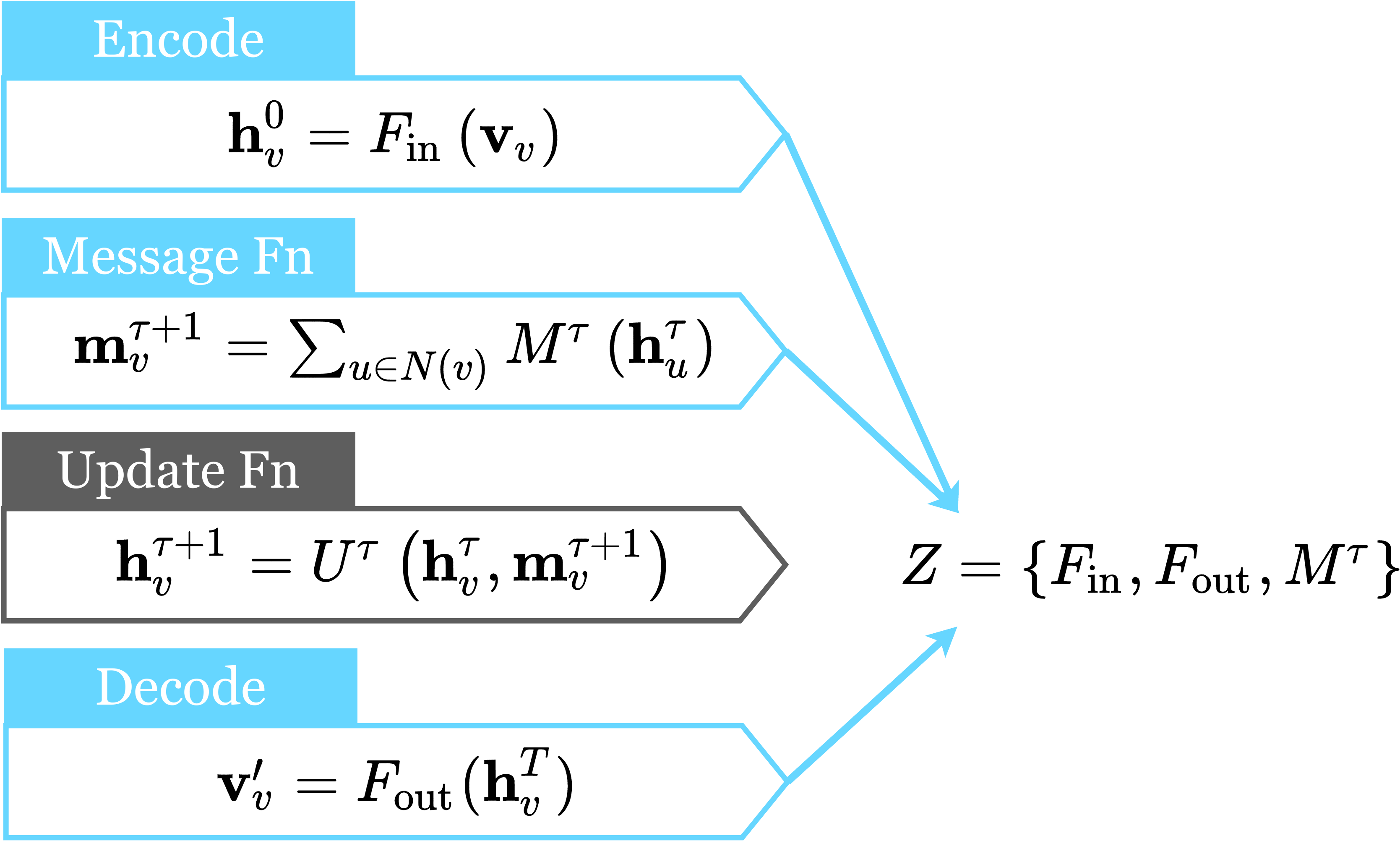}}
\vskip 0.45in
\caption{A visual representation of our \acrshort{sf} algorithm, as outlined in Section \ref{sec:sf}. Prior to training we select a fixed subset $\mathcal{Z} \subseteq \{F^1_{\boldsymbol{\theta}}, \dots, F^n_{\boldsymbol{\theta}}\}$ of the GNN's functions. For our experiments we use $\zeta = \{F_\text{in}, F_\text{out}, M^\tau\}$. Their parameters are then placed in  \acrshort{sf}'s \textit{frozen set} $\zeta = \{\boldsymbol{\theta} \mid F_{\boldsymbol{\theta}} \in Z\} $. During training, \acrshort{sf} excludes parameters in $\zeta$ from being updated by the optimiser.}
\label{fig:sf-diagram-appdx}
\end{center}
\end{figure}

\begin{figure}[h]
\begin{center}
\centerline{\includegraphics[width=0.6\linewidth]{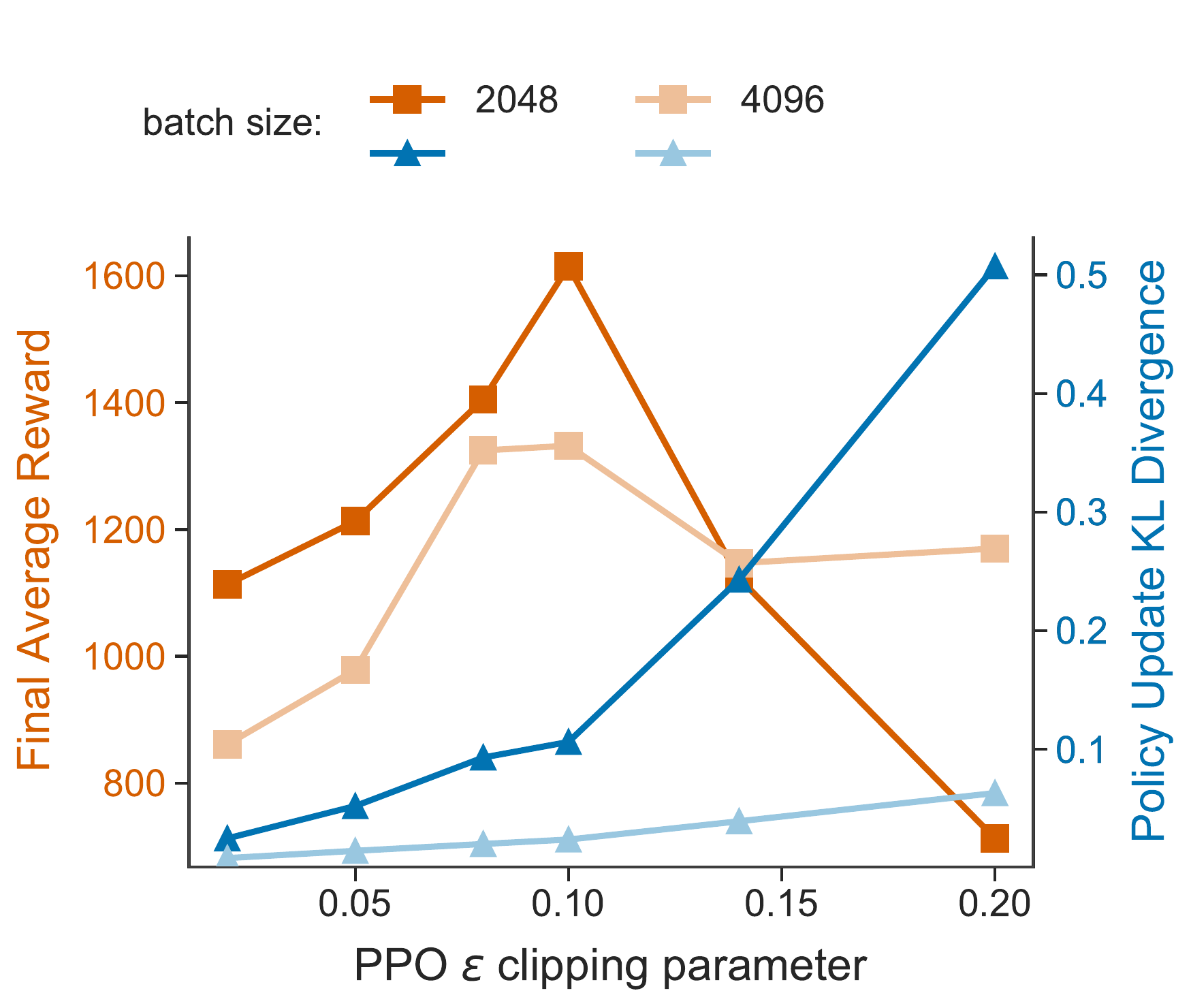}}
\caption{The effect of increasing the batch size on the influence of \acrshort{nn}'s $\epsilon$ clipping hyperparameter (see Figure \ref{fig:kl-clipping}) after ten million timesteps. Increasing the batch size reduces the underlying policy divergence. This makes the algorithm less sensitive to high values of $\epsilon$ (i.e. low clipping), but also leads to a drop in sample efficiency, reducing the maximum reward attained within this time-frame.}
\label{fig:kl-clipping-appdx}
\end{center}
\end{figure}

\begin{figure*}[t]
    \centering
    \includegraphics[width=0.63\linewidth]{figures/sample_efficiency_batch_size.pdf}
    \begin{subfigure}[t]{0.507\linewidth}
        \centering
        \includegraphics[width=\linewidth]{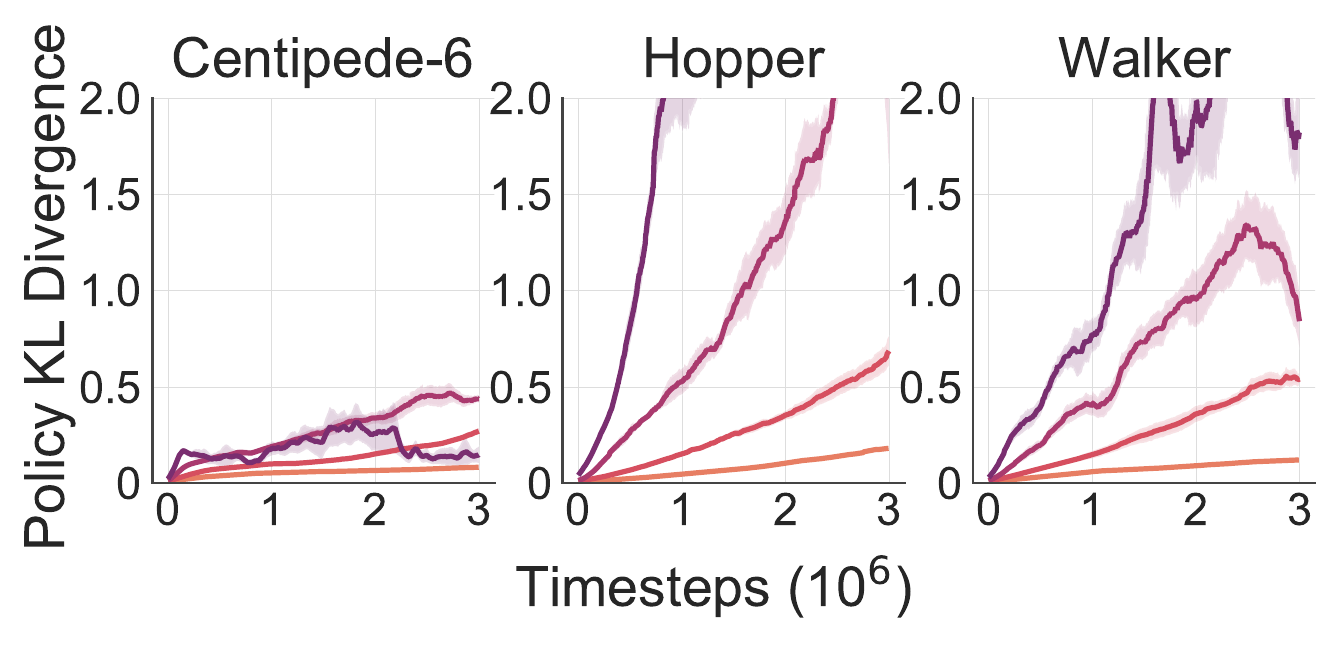}
        \caption{\acrshort{nn}}
    \end{subfigure}%
    \hskip 0.015in
    \vrule
    \hskip 0.015in
    \begin{subfigure}[t]{0.473\linewidth}
        \centering
        \includegraphics[trim={0.8cm 0 0 0}, clip, width=\linewidth]{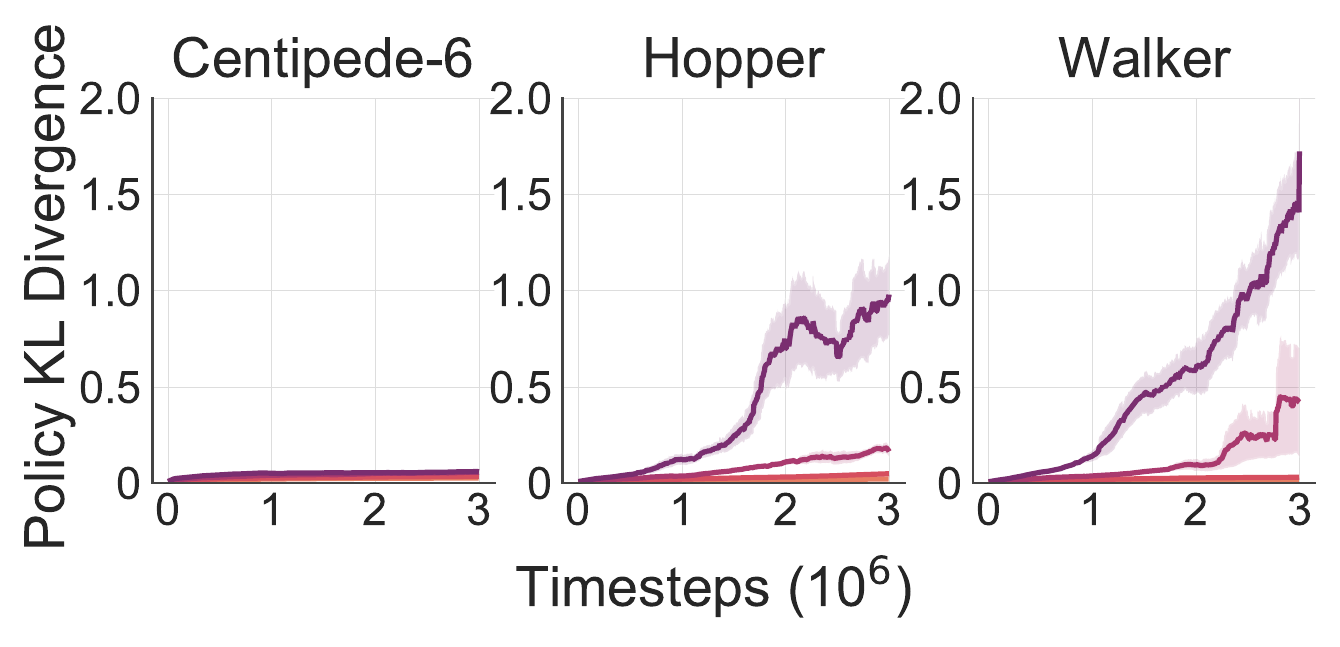}
        \caption{\acrshort{sf}}
    \end{subfigure}
    \caption{Accompanying KL divergence plots for Figure \ref{fig:se-tasks}. As \acrshort{sf} reduces the policy divergence between updates, smaller batch sizes can be used before the KL divergence becomes prohibitively large. This effect underlies the improved sample efficiency demonstrated.}
    \label{fig:se-tasks-apdx}
\end{figure*}

\begin{figure}[h]
\begin{center}
\centerline{\includegraphics[width=0.63\linewidth]{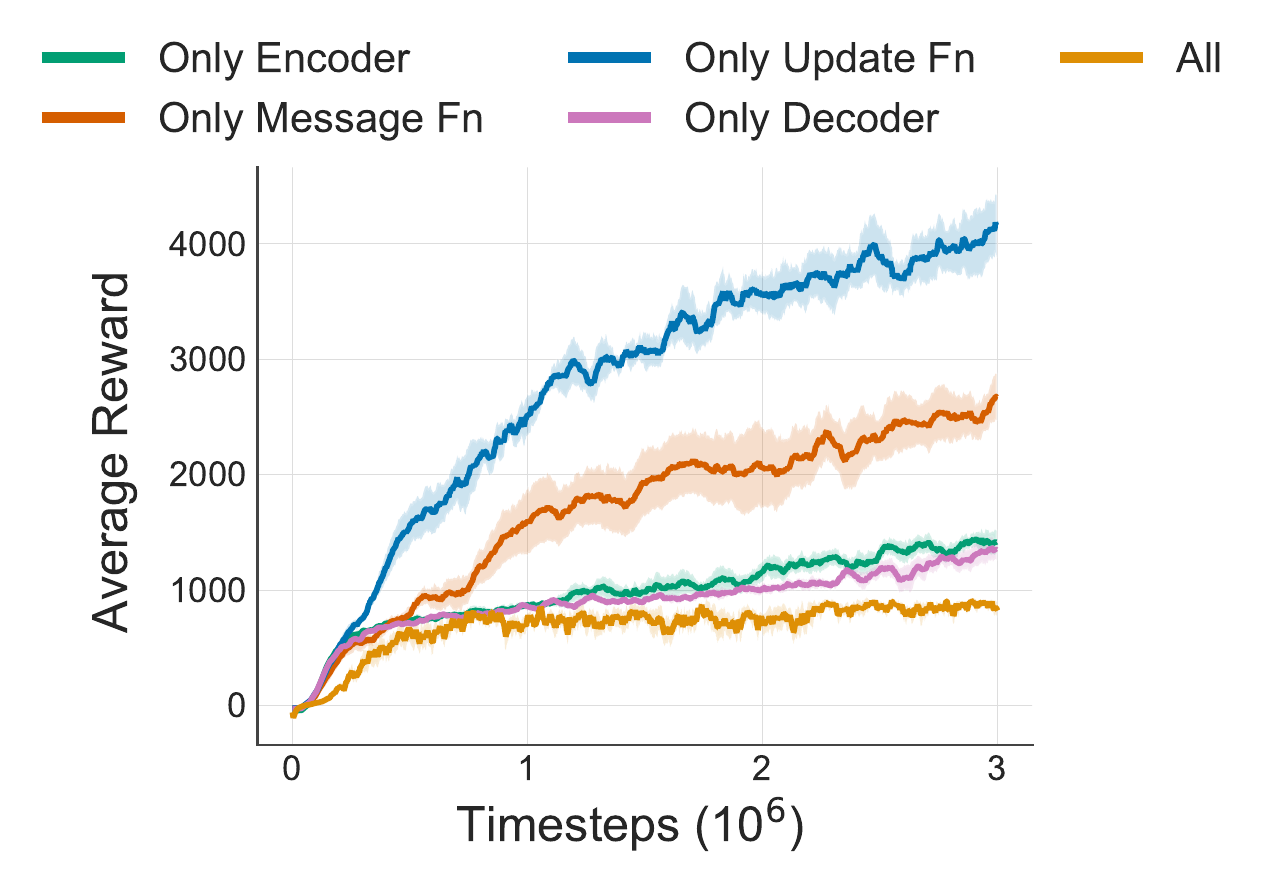}}
\caption{Ablation demonstrating the effect of only training single parts of the network (freezing the rest). The configuration of \acrshort{sf} we use for our experiments is equivalent to only training the update function, which is the most effective approach here, and all approaches are superior to training the entire \acrshort{gnn}. For this experiment, we train on \texttt{Centipede-6} using the small batch size of 256 in all cases. This setting was chosen as it demonstrates the difference in performance for these approaches most clearly.}
\label{fig:ablation}
\end{center}
\vskip -0.2in
\end{figure}